\RequirePackage{graphicx}\RequirePackage{array}
\documentclass{article}

\usepackage[T1]{fontenc}
\usepackage{iclr2027_conference,times}
\usepackage{amsmath,amssymb}
\usepackage{amsmath,amsfonts,bm}

\def\eqref#1{equation~\ref{#1}}

\def\1{\bm{1}}

\DeclareMathAlphabet{\mathsfit}{\encodingdefault}{\sfdefault}{m}{sl}
\SetMathAlphabet{\mathsfit}{bold}{\encodingdefault}{\sfdefault}{bx}{n}

\usepackage{xcolor,colortbl}
\usepackage{booktabs,tabularx,etoc}
\definecolor{storymotionblue}{RGB}{226,240,252}
\newsavebox{\storytablebox}

\newcommand{\storytablefont}{\fontsize{8}{9.4}\selectfont}

\newcommand{\storyendwrap}{
  \par
  \ifnum\value{WF@wrappedlines}>1
    \vspace{\dimexpr\value{WF@wrappedlines}\baselineskip-\baselineskip\relax}
  \fi
  \WFclear
}

\definecolor{storyreviewpurple}{RGB}{126,48,160}
\definecolor{storyyesgreen}{RGB}{25,130,70}
\definecolor{storynored}{RGB}{185,35,45}

\newcommand{\storyyes}{\textcolor{storyyesgreen}{$\checkmark$}}
\newcommand{\storyno}{\textcolor{storynored}{$\times$}}

\definecolor{storyreviewgreen}{RGB}{0,92,62}
\definecolor{storyrevieworange}{RGB}{185,112,0}

\newcommand{\metricstd}[2]{\mbox{#1\textsuperscript{\fontsize{4.5}{5}\selectfont$\pm$#2}}}

\definecolor{storylinkblue}{RGB}{0,82,156}

\newcommand{\acceptededitorial}[1]{#1}

\newcommand{\planbefore}[1]{}

\usepackage{longtable,placeins,capt-of,float}
\usepackage{hyperref}
\hypersetup{hidelinks}
\usepackage{url,wrapfig}

\title{\raggedright AESOP: Asymmetric Human--Camera Generation with Translation-Intensity Control}

\iclrfinalcopy
\author{\begin{minipage}{\textwidth}\raggedright\normalfont
\textbf{Jingzhong Lin\textsuperscript{1,*}, Zhanke Wang\textsuperscript{2}, Heng Li\textsuperscript{3}, Wenxiang Liu\textsuperscript{1}}\\[2pt]
\textbf{Zhao Zhang\textsuperscript{1}, Kecheng Tang\textsuperscript{1}, Dongdong Xiang\textsuperscript{1}, Changbo Wang\textsuperscript{1}}\\[2pt]
\textbf{Di Kang\textsuperscript{4}, Chunchao Guo\textsuperscript{4}, Linchao Bao\textsuperscript{4}, Gaoqi He\textsuperscript{1,\ensuremath{\dagger}}}\\[7pt]
\textsuperscript{1}East China Normal University\\
\textsuperscript{2}Peking University\\
\textsuperscript{3}Sun Yat-sen University\\
\textsuperscript{4}Tencent
\end{minipage}}
\hypersetup{pdftitle={AESOP: Asymmetric Human--Camera Generation with Translation-Intensity Control},pdfauthor={Jingzhong Lin, Zhanke Wang, Heng Li, Wenxiang Liu, Zhao Zhang, Kecheng Tang, Dongdong Xiang, Changbo Wang, Di Kang, Chunchao Guo, Linchao Bao, Gaoqi He}}

\begin{document}

\maketitle
\fancyhead{}\renewcommand{\headrulewidth}{0pt}
\begingroup
\renewcommand{\thefootnote}{\fnsymbol{footnote}}
\footnotetext[1]{This work was completed during an internship at Tencent.}
\footnotetext[2]{Correspondence Author.}
\endgroup

\etocdepthtag{main}
\begin{abstract}
Human motion defines an action, while a camera trajectory determines how it is presented. Camera generation for a given human motion and joint human--camera generation are usually treated as separate tasks, although both share an asymmetric dependency: human motion can be generated independently, whereas the camera responds to the realized action. We introduce \mbox{\textbf{AESOP}}, a unified framework with an independent human pathway and a shared human-conditioned camera generator. Its asymmetric architecture serves both tasks while preserving the human output during camera generation.
Although human context anchors the shot to the action and camera text describes its movement, translation intensity remains underspecified. We therefore construct trajectory pairs that differ in camera translation magnitude while sharing human motion and camera text, then use these pairs to learn an explicit intensity condition. Experiments on the PulpMotion dataset demonstrate strong camera distributional and framing quality in both tasks and effective control over camera translation intensity.
Project page is available at \url{https://ripemangobox.github.io/AESOP/}.
\end{abstract}

\section{Introduction}
\label{sec:introduction}

\suppressfloats[t]
\begin{figure*}[t]
    \centering
    \includegraphics[width=0.99\textwidth]{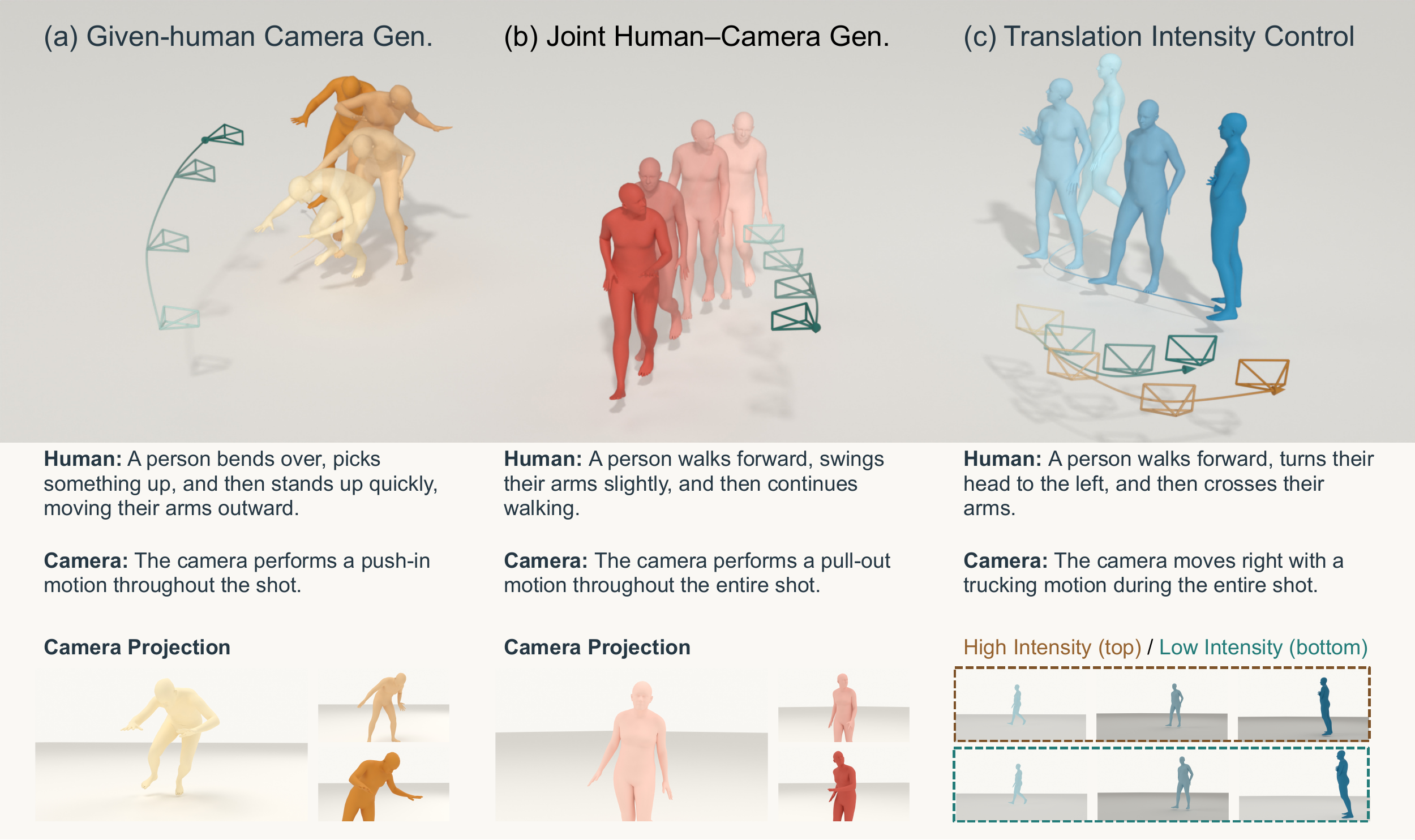}
\caption{AESOP supports (a) camera generation for a given human motion, (b) joint human--camera generation, and (c) continuous camera translation-intensity control (the amount of camera travel) in both tasks without changing the supplied or generated human motion.}
\label{fig:teaser}
\end{figure*}

Text-conditioned human motion generation can now produce diverse, natural actions that follow language descriptions~\citep{guo2022humanml3d,tevet2023mdm,zhang2023motiondiffuse,chen2023mld,zhang2023t2mgpt,guo2024momask}. Authoring an animation also requires choosing how the action is viewed. Cinematography treats framing and camera movement as constructive choices that organize screen space, direct attention, and shape how viewers experience an event~\citep{bordwell2023filmart,brown2022cinematography}.
Human motion and camera trajectories therefore play complementary but not interchangeable roles: human motion defines the action, whereas camera trajectory presents that action for the viewer. A human-aware camera is therefore part of motion storytelling.

Prior work approaches this relationship through two broad formulations. Conditional camera methods adapt a trajectory or shot plan to an available human motion~\citep{courant2024et,kizil2026auteur}, leaving human synthesis outside the task. Coupled systems synthesize human motion and camera trajectories within one generative process~\citep{courant2026pulpmotion,cheng2026storytelling}, treating their generation as interdependent. These formulations leave an opportunity to unify both tasks around the distinct roles of human action and camera response.

We introduce \textbf{AESOP} (\textbf{A}symmetric \textbf{E}ncoding and \textbf{S}ynthesis of \textbf{O}bserver \textbf{P}aths). Two observations guide our design. First, human action can be determined independently of the camera, while both tasks can share a camera generator conditioned on supplied or generated human motion. Second, human context and camera text specify an action and a shot description but leave translation intensity underspecified. Together, these observations motivate a shared asymmetric architecture and explicit geometric supervision for camera control.

For the architecture, we propose an independent human representation and generator together with a shared human-conditioned camera generator through a common latent interface. Camera training freezes the human pathway; joint inference completes human generation before sampling the camera trajectory. This one-way dependency serves both tasks while preserving the supplied or generated human action during camera generation.

To learn explicit translation intensity, we construct \emph{intensity-paired augmentation} (IPA): weaker and stronger camera trajectories which share human motion and camera text. IPA contracts or expands smooth displacement about the camera center at the first frame, and geometry and visibility checks screen the targets. Auxiliary \emph{direction-paired augmentation} (DPA) supplies opposite-direction Truck and Dolly examples. After original-pair camera training, DPA and then IPA are introduced through continuation while retaining earlier data.

On the PulpMotion dataset, AESOP achieves strong camera distributional and framing quality in both tasks. It reduces given-human r-FPD by 49.9\% relative to DIRECTOR-C and joint camera FDC by 87.8\% relative to PulpMotion DiT. \acceptededitorial{Its learned intensity condition controls camera travel.}

Our contributions are threefold. (1) We introduce a unified asymmetric framework with an independent human pathway and a shared human-conditioned camera generator for given-human and joint generation. (2) We learn explicit camera translation-intensity control from geometry-grounded intensity pairs, with directional pairs providing auxiliary supervision. (3) We evaluate both tasks, demonstrating substantially improved camera distributional and framing quality, strong human distribution quality, and effective camera travel intensity control.

\section{Related Work}
\label{sec:related-work}

\subsection{Human motion generation}

Text-to-motion research has expanded from synthesizing plausible action clips to supporting more detailed and flexible authoring. Paired motion--language data such as HumanML3D~\citep{guo2022humanml3d} and MotionMillion~\citep{fan2025motionmillion} made semantic fidelity and motion quality central evaluation targets. Diffusion-based models~\citep{tevet2023mdm,zhang2023motiondiffuse,chen2023mld} and discrete generative models~\citep{zhang2023t2mgpt,guo2024momask} offer complementary approaches, with masked autoregressive diffusion extending continuous motion modeling~\citep{add_meng2025mardm}. As clip synthesis improved, the focus broadened to instructions and controls beyond a single action label: finer supervision addresses spatial and temporal detail~\citep{wu2025finemotion}, streaming generation accommodates incoming instructions~\citep{smadd_zhao2025dart,xiao2025motionstreamer}, and larger-scale training expands action coverage and controllability~\citep{wen2025hymotion,rempe2026kimodo}. Video-generation priors also support motion synthesis~\citep{add_lin2026vimogen}.

\subsection{Human-aware camera generation}

Camera synthesis accounts for the subject it depicts. For an available performance, the problem is to generate a compatible viewpoint and trajectory: DIRECTOR~\citep{courant2024et} introduces character-aware text-to-camera generation, while DanceCamera3D~\citep{wang2024dancecamera3d} studies camera movement conditioned on dance and music. This line of work establishes human motion as useful camera context. When the performance itself is also generated, fixing human motion as input is insufficient. PulpMotion~\citep{courant2026pulpmotion} incorporates screen-space framing into human--camera generation, and Towards Storytelling Animations~\citep{cheng2026storytelling} studies joint synthesis of their temporal behavior. \acceptededitorial{AESOP uses a shared complete-sequence human interface to support both supplied and generated performances.}

\subsection{Explicit and geometry-grounded camera control}

Geometric camera interfaces support composition constraints~\citep{smadd_lino2015toric}; learned keyframing systems combine camera styles with keyframe and velocity control~\citep{add_jiang2021keyframing}. GenDoP generates camera trajectories from text, optionally conditioned on initial-frame RGB-D observations~\citep{zhang2025gendop}. Actor-relative framing in Auteur~\citep{kizil2026auteur} and visual preference optimization in VERTIGO~\citep{li2026vertigo} improve how directing intent is expressed or followed. Even with human context, however, one instruction can admit many trajectories, leaving the strength and direction of movement ambiguous. Explicit trajectory interfaces address this ambiguity by supplying the geometry itself: MotionCtrl~\citep{wang2024motionctrl} and CameraCtrl~\citep{he2024cameractrl} condition video generation on camera paths. This shifts trajectory authoring to the user. AESOP instead supplies training examples that vary direction or translation intensity while holding human motion fixed. Paired directional instructions and a continuous intensity scalar expose these geometric choices without requiring a target path.

\section{Method}
\label{sec:method}

\subsection{Preliminaries}
\label{sec:preliminaries}

\paragraph{Problem definition.}
Let $H=(h_0,\ldots,h_{T-1})$ denote human motion and $C=(c_0,\ldots,c_{T-1})$ a synchronized camera trajectory, with text instructions $T_H$ and $T_C$. Following PulpMotion~\citep{courant2026pulpmotion}, human motion data encode root motion and heading, local joint positions, and joint rotations; camera trajectory data encode human-relative position, camera rotation, field of view, and translation velocity. \textbf{Given-human camera generation} maps $(H,T_C)$ to $C$. \textbf{Joint human--camera generation} maps $(T_H,T_C)$ to $(H,C)$. Both tasks accept an optional translation-intensity condition $a$, with $a=1$ specifying default behavior and smaller or larger values requesting weaker or stronger camera-center translation.

\paragraph{Flow matching for human and camera generation.}
\acceptededitorial{For flow time $\sigma\in[0,1]$, clean latent $z$ and Gaussian noise $\epsilon$, define} $z_\sigma=(1-\sigma)z+\sigma\epsilon$ and target velocity $\epsilon-z$~\citep{smadd_lipman2023fm}. The human and camera flows minimize
\begin{equation}
\begin{aligned}
    \mathcal L_H^{\mathrm{FM}}
    &=\mathbb E\!\left[\left\|v_\phi(z_{H,\sigma},\sigma,T_H)-(\epsilon_H-z_H)\right\|_2^2\right],\\
    \mathcal L_C^{\mathrm{FM}}
    &=\mathbb E\!\left[\left\|v_\theta(z_{C,\sigma},\sigma,z_H,T_C,a)-(\epsilon_C-z_C)\right\|_2^2\right].
\end{aligned}
\label{eq:human-camera-flow-matching}
\end{equation}
Here $z_H$ and $z_C$ are normalized human and camera latents, and the camera flow receives the complete, fixed human latent sequence. Sampling integrates the predicted velocity from noise at $\sigma=1$ to a clean latent at $\sigma=0$.

\subsection{Unified asymmetric human--camera generation}

\acceptededitorial{AESOP learns its asymmetric representation in Stage~1 and the human and camera generators in Stage~2. Geometry-grounded pairs provide translation-intensity and auxiliary direction supervision.}

\suppressfloats[t]
\begin{figure}[!htb]
    \centering
    \input{sections/figures/pipeline}
\caption{\acceptededitorial{AESOP overview. (A) Separate encoders produce 128-channel human and 64-channel camera latents; the camera decoder reads their concatenation. (B,C) Human generation supplies the complete latent sequence to both human decoding and camera conditioning. Given-human generation instead encodes the \acceptededitorial{supplied motion}. (D,E) Human and camera Transformer blocks.}}
\label{fig:pipeline}
\end{figure}

As illustrated in Fig.~\ref{fig:pipeline}(B,C), AESOP factorizes generation as
\begin{equation}
    p(H,C\mid T_H,T_C,a)=p_\phi(H\mid T_H)\,p_\theta(C\mid H,T_C,a).
    \label{eq:asymmetric-factorization}
\end{equation}
\acceptededitorial{Both tasks use the same conditional camera model $p_\theta(C\mid H,T_C,a)$. Given-human generation encodes the supplied motion, whereas joint generation first samples the complete human latent from $p_\phi(H\mid T_H)$ and reuses it for camera conditioning and human decoding.}

\paragraph{Asymmetric human--camera representation.}
Stage~1 (Fig.~\ref{fig:pipeline}(A)) first trains a human encoder--decoder and then freezes it while training the camera encoder--decoder:
\begin{equation}
\begin{aligned}
    z_H&=E_H(H), & \hat H&=D_H(z_H),\\
    z_C&=E_C(C), & \hat C&=D_C\!\left([\operatorname{sg}(z_H);z_C]\right).
\end{aligned}
\label{eq:asymmetric-representation}
\end{equation}

Here $[\cdot;\cdot]$ denotes channel concatenation and $\operatorname{sg}$ stops gradients. The camera encoder maps camera trajectory data to a latent sequence, and the camera decoder combines this sequence with the frozen human latents. \acceptededitorial{Human reconstruction depends only on $z_H$.} Supplementary Section~\ref{supp:interface} illustrates the difference from PulpMotion's symmetric representation. Stage~1 uses reconstruction and temporal-difference losses, with additional human root and yaw terms; the full objectives appear in Supplementary Section~\ref{supp:training}.

\paragraph{\acceptededitorial{Independent human generation.}}
\acceptededitorial{In Stage~2, we train the human flow (Fig.~\ref{fig:pipeline}(B,D)) with the objective in Eq.~\ref{eq:human-camera-flow-matching}, then freeze its parameters before camera training. At inference, the human flow samples a complete latent sequence from human text and Gaussian noise, and the frozen human decoder reconstructs the motion. For fixed human text and noise, camera guidance, intensity and camera noise leave the generated human motion unchanged.}

\paragraph{Human-aware camera generation.}
During training, the camera flow (Fig.~\ref{fig:pipeline}(C,E)) minimizes $\mathcal L_C^{\mathrm{FM}}$ over $\theta$, \acceptededitorial{conditioned on the complete 128-channel encoded ground-truth human sequence while \acceptededitorial{predicting velocity in the 64-channel camera latent space}. The human flow and Stage~1 representation remain frozen.} Training has three phases within Stage~2: original-pair training, DPA continuation, and IPA continuation. Both continuation phases mix original and augmented pairs at predefined sampling ratios, with DPA pairs retained during IPA training. Table~\ref{tab:optimization} gives the schedule.

At inference, given-human generation uses the encoded supplied motion, while joint generation uses the completed human-flow latent directly (Fig.~\ref{fig:pipeline}(B,C)). The same camera flow and decoder serve both tasks. Camera training drops text with probability 0.1 while retaining human context and intensity, enabling classifier-free guidance~\citep{smadd_ho2022cfg} between the text-conditioned velocity $v_\theta^c$ and text-dropped velocity $v_\theta^u$:
\begin{equation}
    v_\theta^{g}=v_\theta^u+g\bigl(v_\theta^c-v_\theta^u\bigr).
    \label{eq:camera-cfg}
\end{equation}

The text-conditioned and text-dropped velocity predictions use the same human latent, intensity $a$, noisy camera latent, and flow time. \acceptededitorial{Here $g$ is the camera-text guidance weight.}

\subsection{Geometry-grounded camera control}

Ordinary paired data rarely show how the same action should be filmed when only camera direction or translation intensity changes. DPA constructs opposite-direction targets, while IPA constructs targets at different translation intensities in physical camera space (Fig.~\ref{fig:geometry-programs}).

\begin{figure}[!htbp]
\centering
\input{sections/figures/control_pairs}
\caption{Training target construction of DPA and IPA. (a,b) Opposite directions share human motion and the initial camera state. (c,d) Intensity variants retain human motion and camera text.}
\label{fig:geometry-programs}
\vspace{-10pt}
\end{figure}

\paragraph{Data construction.}

We mine high-confidence atomic Truck and Dolly events for opposite-direction pairs, and translation-active trajectories for weaker/stronger intensity variants. Both constructions fix human motion. Geometry, visibility, framing, dynamics and representation checks screen the targets; Supplementary Section~\ref{supp:data} details the construction and acceptance criteria for both DPA and IPA.

Frames are indexed by $t=0,\ldots,T-1$. A direction event occupies $[t_s,t_e]\subseteq[0,T-1]$ which may cover a subinterval of the sequence. For direction pairing, let $q^+(t)$ be the original signed event coordinate over $[t_s,t_e]$: lateral camera-center position for trucking, or camera--human radial distance for dolly translation. The opposite target $q^-(t)$ reflects its displacement about the event start. For intensity control, we smooth the original world-space camera-center path $p_t$ to obtain its low-frequency component $s_t$; $r_t=p_t-s_t$ contains the remaining local fluctuations. The two transformations are
\begin{equation}
\begin{aligned}
    q^-(t)&=2q^+(t_s)-q^+(t),\qquad t\in[t_s,t_e],\\
    p_t^{(a)}&=p_0+a(s_t-s_0)+\min(a,1)(r_t-r_0).
\end{aligned}
\label{eq:geometry-programs}
\end{equation}
Direction targets (Fig.~\ref{fig:geometry-programs}\hyperref[fig:direction-original]{a}--\hyperref[fig:direction-reversed]{b}) use paired minimal text templates that differ only in direction and preserve event timing, the displacement magnitude of $q$, initial camera state, rotation, and field of view; an endpoint offset maintains continuity after the event. Constructed intensity targets (Fig.~\ref{fig:geometry-programs}\hyperref[fig:intensity-reduced]{c},\hyperref[fig:intensity-increased]{d}) preserve camera text, rotation, field of view, and sequence timing. They recover $p_t$ exactly at $a=1$ and amplify only the smooth component when $a>1$, avoiding amplification of residual jitter.

 We also select low-translation sources and pair each unchanged camera target with both $a=1$ and a non-default intensity. These unchanged-target pairs enter the final training phase alongside active variants, teaching the model to retain low-translation shots under intensity changes. We embed intensity with a two-layer MLP, $e_a=W_2\operatorname{SiLU}(W_1[a-1,(a-1)^2]^\top)$, and add $e_a$ to the camera flow's timestep condition (Fig.~\ref{fig:pipeline}(C)). The learned projections $W_1,W_2$ are bias-free and $W_2$ is zero-initialized, giving $e_1=0$ throughout training and an initially inactive intensity pathway.

\paragraph{Paired supervision.}
\acceptededitorial{Both members of each control pair contribute to the flow-matching objective in Eq.~\ref{eq:human-camera-flow-matching}, sharing human context, Gaussian noise, flow time and camera-text dropout decisions. DPA changes the directional prompt with the target; IPA retains the prompt and changes the intensity label. \acceptededitorial{\acceptededitorial{During target construction, candidate intensities are sampled in $0<a<1$ and $1<a<2$, with values nearer $a=1$ proposed more often. Each retained translation-active source supplies one accepted weaker target and one accepted stronger target.}}}

\storyendwrap

\raggedbottom
\begingroup
\setlength{\textfloatsep}{10pt plus 2pt minus 2pt}
\setlength{\floatsep}{14pt plus 2pt minus 2pt}
\setlength{\intextsep}{10pt plus 2pt minus 2pt}
\setcounter{topnumber}{3}
\setcounter{bottomnumber}{2}
\setcounter{totalnumber}{5}
\renewcommand{\topfraction}{0.95}
\renewcommand{\bottomfraction}{0.9}
\renewcommand{\textfraction}{0}
\renewcommand{\floatpagefraction}{0.85}
\section{Experiments}

\label{sec:experiments}
We evaluate camera quality in given-human and joint human--camera generation, human generation quality, and geometry-grounded camera control.

\subsection{Experimental setup}\label{sec:setup}

\paragraph{Data and implementation.}
\acceptededitorial{We use the public PulpMotion split~\citep{courant2026pulpmotion}, with 162,760 training sequences and 4,053 test sequences, evaluating both tasks at valid sequence lengths. Geometry-grounded training uses 601 direction sources, 8,000 translation-active sources and 2,000 low-translation sources from the training split. Unless stated otherwise, AESOP uses \acceptededitorial{50-step Euler sampling per stream}, human guidance 1, camera guidance $g=1.5$ and translation intensity $a=1$. Both tasks share the camera model. Supplementary Sections~\ref{supp:interface} and~\ref{supp:training} give architecture and optimization details.}

\paragraph{Metrics.}
\acceptededitorial{Table~\ref{tab:main} reports camera distribution distance (FDC), camera-text alignment (CLaTr), movement-caption F1~\citep{courant2024et}, and PRDC recall in the frozen CLaTr embedding space~\citep{naeem2020prdc}. Framing is measured by r-FPD, the distribution distance of projected human joints in normalized image coordinates, and Out (\%), the valid-frame rate at which none of nine key joints is in front of the camera and inside the image~\citep{courant2026pulpmotion}. \acceptededitorial{Camera-center ADE measures trajectory error (m) for reconstruction and given-human generation;} rotation error ($^\circ$) compares orientations with the reference camera, including in joint generation. Human metrics use a common frozen evaluator~\citep{smadd_petrovich2023tmr}: distribution distance $\mathrm{FD}_{\mathrm{TMR}}$, text alignment TMR and PRDC. CLaTr and TMR report mean nonnegative text cosine similarity scaled by 100. Table~\ref{tab:ae-reconstruction} adds reconstruction diagnostics. }

\acceptededitorial{Repeated-run results use $\mathrm{mean}^{\pm\mathrm{SD}}$\acceptededitorial{, where SD denotes sample standard deviation}. Generation and reconstruction comparisons aggregate three independent training seeds; direction-pair results aggregate three paired camera-noise sets from one training seed. Control sweeps fix one checkpoint and share sampling noise across settings.}

\begin{table}[!tp]
\setlength{\abovecaptionskip}{0pt}\setlength{\belowcaptionskip}{6pt}
\caption{\acceptededitorial{\acceptededitorial{Quantitative comparison.} \acceptededitorial{AE rows report reconstruction; the remaining model rows report generation.} AESOP (no aug.) uses original pairs, (+DPA) adds direction pairs, and full AESOP further adds intensity pairs. Human scores are shared across AESOP camera-training phases. Bold marks the best displayed value within each reconstruction/task group, excluding GT; blue marks AESOP.}}

\label{tab:main}

\centering\begingroup\fontsize{7}{8.4}\selectfont
\setlength{\tabcolsep}{0.6pt}\renewcommand{\arraystretch}{1.15}
\begin{tabular*}{\textwidth}{@{\extracolsep{\fill}}l*{10}{r}@{}}
\toprule
 & \multicolumn{4}{c}{camera distribution \& alignment} & \multicolumn{4}{c}{camera framing \& geometry} & \multicolumn{2}{c}{\acceptededitorial{\shortstack{human distribution\\\& alignment}}}\\
\cmidrule(lr){2-5}\cmidrule(lr){6-9}\cmidrule(lr){10-11}
Method & FDC$\downarrow$ & CLaTr$\uparrow$ & F1$\uparrow$ & Recall$_C$$\uparrow$ & r-FPD$\downarrow$ & Out$\downarrow$ & ADE$\downarrow$ & Rot.$\downarrow$ & FD$_{\mathrm{TMR}}$$\downarrow$ & TMR$\uparrow$\\\midrule
GT reference & $\approx 0$ & 70.237 & 0.945 & 1.00 & 0.003 & 0.71 & 0.000 & 0.002 & $\approx 0$ & 18.398 \\
PulpMotion AE (sym.) & \metricstd{16.59}{3.52} & \metricstd{59.91}{1.38} & \metricstd{0.739}{0.032} & \metricstd{0.973}{0.005} & \metricstd{0.133}{0.02} & \metricstd{3.93}{0.55} & \metricstd{0.110}{0.01} & \metricstd{1.37}{0.10} & \metricstd{38.79}{9.99} & \textbf{\metricstd{20.15}{0.28}}\\
\rowcolor{storymotionblue}
AESOP AE (asym.) & \textbf{\metricstd{0.11}{0.10}} & \textbf{\metricstd{70.10}{0.13}} & \textbf{\metricstd{0.941}{0.001}} & \textbf{\metricstd{0.999}{0.0004}} & \textbf{\metricstd{0.089}{0.01}} & \textbf{\metricstd{2.93}{0.28}} & \textbf{\metricstd{0.029}{0.0006}} & \textbf{\metricstd{0.41}{0.01}} & \textbf{\metricstd{10.05}{0.89}} & \metricstd{17.56}{0.12}\\
\midrule\multicolumn{11}{l}{\textit{Given-human camera generation}}\\\midrule
DanceCamera3D & \metricstd{180.08}{5.24} & \metricstd{20.92}{0.54} & \metricstd{0.252}{0.002} & \metricstd{0.600}{0.01} & \metricstd{2.692}{0.48} & \metricstd{21.72}{2.33} & \metricstd{2.995}{0.07} & \metricstd{63.52}{1.19} & \multicolumn{1}{c}{--} & \multicolumn{1}{c@{}}{--} \\
CCD & \metricstd{359.70}{6.02} & \metricstd{9.53}{0.19} & \metricstd{0.118}{0.004} & \metricstd{0.632}{0.02} & \metricstd{1.937}{0.19} & \metricstd{0.63}{0.10} & \metricstd{2.891}{0.04} & \metricstd{75.58}{2.09} & \multicolumn{1}{c}{--} & \multicolumn{1}{c@{}}{--} \\
CCD-H & \metricstd{375.97}{39.43} & \metricstd{11.32}{1.58} & \metricstd{0.134}{0.020} & \metricstd{0.706}{0.06} & \metricstd{2.942}{2.35} & \textbf{\metricstd{0.51}{0.04}} & \metricstd{2.712}{0.14} & \metricstd{68.49}{2.60} & \multicolumn{1}{c}{--} & \multicolumn{1}{c@{}}{--} \\
DIRECTOR-C & \metricstd{7.29}{0.42} & \metricstd{64.33}{0.28} & \metricstd{0.848}{0.001} & \textbf{\metricstd{0.897}{0.003}} & \metricstd{1.993}{0.04} & \metricstd{10.26}{0.22} & \metricstd{2.687}{0.07} & \metricstd{58.16}{0.36} & \multicolumn{1}{c}{--} & \multicolumn{1}{c@{}}{--} \\
AESOP (no aug.) & \metricstd{6.99}{0.20} & \metricstd{69.60}{0.33} & \metricstd{0.905}{0.006} & \metricstd{0.866}{0.01} & \metricstd{1.160}{0.07} & \metricstd{12.56}{0.43} & \metricstd{1.577}{0.01} & \metricstd{33.00}{0.46} & \multicolumn{1}{c}{--} & \multicolumn{1}{c@{}}{--} \\
AESOP (+ DPA) & \metricstd{6.03}{0.17} & \metricstd{70.65}{0.28} & \metricstd{0.915}{0.004} & \metricstd{0.860}{0.004} & \metricstd{1.001}{0.03} & \metricstd{11.51}{0.21} & \metricstd{1.515}{0.01} & \metricstd{32.10}{0.36} & \multicolumn{1}{c}{--} & \multicolumn{1}{c@{}}{--} \\
\rowcolor{storymotionblue}
AESOP & \textbf{\metricstd{5.97}{0.16}} & \textbf{\metricstd{70.72}{0.32}} & \textbf{\metricstd{0.916}{0.005}} & \metricstd{0.863}{0.01} & \textbf{\metricstd{0.998}{0.04}} & \metricstd{11.47}{0.22} & \textbf{\metricstd{1.512}{0.01}} & \textbf{\metricstd{32.06}{0.41}} & \multicolumn{1}{c}{--} & \multicolumn{1}{c@{}}{--} \\
\midrule\multicolumn{11}{l}{
\textit{Joint human--camera generation}}\\\midrule
PulpMotion DiT & \metricstd{83.80}{3.30} & \metricstd{44.92}{2.36} & \metricstd{0.592}{0.027} & \metricstd{0.711}{0.02} & \metricstd{7.772}{0.73} & \metricstd{35.39}{1.28} & \multicolumn{1}{c}{--} & \textbf{\metricstd{66.76}{1.82}} & \metricstd{366.21}{21.52} & \textbf{\metricstd{24.20}{0.97}} \\
PulpMotion MAR & \metricstd{131.55}{9.14} & \metricstd{38.64}{2.42} & \metricstd{0.514}{0.033} & \metricstd{0.769}{0.03} & \metricstd{8.169}{0.65} & \metricstd{39.08}{2.14} & \multicolumn{1}{c}{--} & \metricstd{70.53}{1.45} & \metricstd{319.98}{7.38} & \metricstd{21.57}{0.87} \\
AESOP (no aug.) & \metricstd{11.39}{0.67} & \metricstd{68.94}{0.34} & \metricstd{0.880}{0.002} & \textbf{\metricstd{0.826}{0.01}} & \metricstd{0.665}{0.03} & \metricstd{9.13}{0.24} & \multicolumn{1}{c}{--} & \metricstd{71.56}{0.53} & \textbf{\metricstd{100.98}{1.91}} & \metricstd{20.46}{0.34} \\
AESOP (+ DPA) & \metricstd{10.38}{0.47} & \metricstd{69.76}{0.33} & \metricstd{0.887}{0.005} & \metricstd{0.824}{0.01} & \textbf{\metricstd{0.576}{0.02}} & \textbf{\metricstd{8.45}{0.08}} & \multicolumn{1}{c}{--} & \metricstd{71.82}{0.88} & \textbf{\metricstd{100.98}{1.91}} & \metricstd{20.46}{0.34} \\
\rowcolor{storymotionblue}
AESOP & \textbf{\metricstd{10.18}{0.50}} & \textbf{\metricstd{69.81}{0.34}} & \textbf{\metricstd{0.891}{0.006}} & \metricstd{0.820}{0.01} & \metricstd{0.581}{0.02} & \metricstd{8.47}{0.04} & \multicolumn{1}{c}{--} & \metricstd{71.87}{0.85} & \textbf{\metricstd{100.98}{1.91}} & \metricstd{20.46}{0.34} \\
\bottomrule
\end{tabular*}\endgroup

\end{table}

\newcommand{\aesopqualitativecomparison}{
\par\smallskip\raggedright\noindent\textbf{Qualitative comparison.} \acceptededitorial{Figure~\ref{fig:qualitative-comparison} shows AESOP following the pull-out prompt while retaining the subject, whereas DIRECTOR-C and DanceCamera3D crop the subject and CCD misses the requested movement. In the joint example, AESOP progresses from full-body to upper-body framing; DiT alternates between cropped and empty views, while MAR largely misses the human.} \acceptededitorial{Supplementary Section~\ref{supp:more-results} provides more comparisons; Section~\ref{supp:ood-humanml3d} shows external examples from HumanML3D and HY-Motion.}\par
}
\begin{figure}[!tp]
\centering
\includegraphics[width=\linewidth]{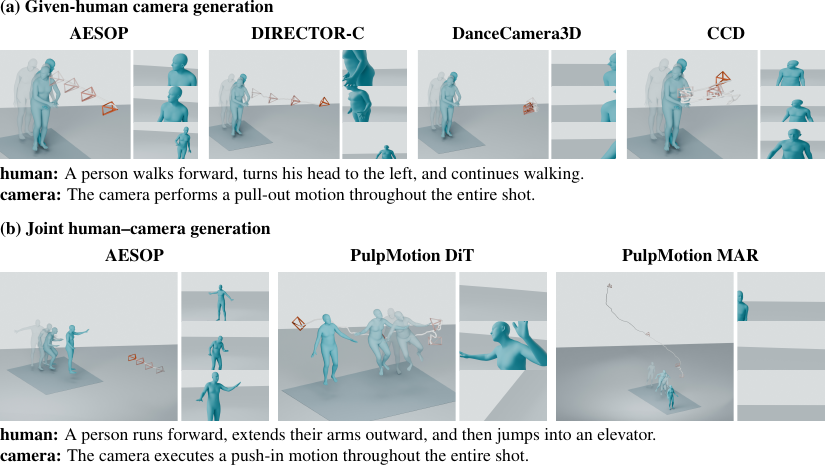}
\caption{\textbf{Qualitative comparison on test examples.} (a) Given-human camera generation with a pull-out prompt. (b) Joint human--camera generation with a push-in prompt. Each method has a global view and three selected camera projections, ordered from top to bottom in time. \acceptededitorial{Global views show human--camera geometry at individually fitted scales; projections show framing over time.}}
\label{fig:qualitative-comparison}
\aesopqualitativecomparison
\end{figure}

\subsection{Comparison with prior methods}

\paragraph{Baselines.}
Given-human comparisons include DIRECTOR-C~\citep{courant2024et}, \acceptededitorial{DanceCamera3D}~\citep{wang2024dancecamera3d}, CCD~\citep{jiang2024ccd}, and our CCD-H adaptation. DIRECTOR-C conditions on camera text and the human root-translation trajectory; DanceCamera3D replaces music with camera text. CCD and CCD-H share a human-relative camera representation. CCD uses a text-only network; CCD-H additionally reads human joints and replaces the text prefix with text cross-attention. PulpMotion~\citep{courant2026pulpmotion} uses DiT~\citep{add_peebles2023dit} and MAR~\citep{add_li2024mar} backbones for joint human--camera generation. \acceptededitorial{Table~\ref{tab:main} uses original-data endpoints for all baselines.} Supplementary Section~\ref{supp:baseline-protocols} details adaptations and training exposure.

\paragraph{Quantitative comparison.}
\acceptededitorial{AESOP leads both tasks in FDC, CLaTr and caption F1 (Table~\ref{tab:main}). Given-human FDC drops from DIRECTOR-C's 7.29 to 5.97; joint FDC drops from PulpMotion DiT's 83.80 to 10.18, while CLaTr rises from 44.92 to 69.81. Without augmentation, AESOP attains FDC 6.99/11.39 for given-human/joint generation, below DIRECTOR-C and PulpMotion DiT, respectively. DPA further improves camera quality; IPA adds intensity control while largely preserving generation quality. For framing, AESOP reduces r-FPD from DIRECTOR-C's 1.993 to 0.998 in given-human generation and from DiT's 7.772 to 0.581 in joint generation. Human $\mathrm{FD}_{\mathrm{TMR}}$ is 100.98 versus DiT/MAR's 366.21/319.98, with lower TMR alignment. CCD-H achieves the lowest given-human Out.}

\begin{table}[!htbp]
\setlength{\abovecaptionskip}{0pt}\setlength{\belowcaptionskip}{6pt}
\caption{\acceptededitorial{\acceptededitorial{Human distributional fidelity and coverage in joint generation in the common frozen human embedding space.} Bold marks the best value in each column.}}
\label{tab:distribution-details}

\centering\begingroup\storytablefont
\setlength{\tabcolsep}{3pt}\renewcommand{\arraystretch}{1.15}
\begin{tabularx}{\textwidth}{@{}l*{4}{>{\raggedleft\arraybackslash}X}@{}}
\toprule
Method & Precision $\uparrow$ & Recall $\uparrow$ & Density $\uparrow$ & Coverage $\uparrow$ \\\midrule
PulpMotion DiT & \metricstd{0.809}{0.02} & \metricstd{0.215}{0.02} & \metricstd{0.748}{0.05} & \metricstd{0.465}{0.02} \\
PulpMotion MAR & \metricstd{0.802}{0.005} & \metricstd{0.312}{0.01} & \metricstd{0.705}{0.02} & \metricstd{0.504}{0.002} \\
\rowcolor{storymotionblue} AESOP & \textbf{\metricstd{0.879}{0.002}} & \textbf{\metricstd{0.835}{0.01}} & \textbf{\metricstd{0.933}{0.02}} & \textbf{\metricstd{0.785}{0.01}} \\
\bottomrule\end{tabularx}\endgroup

\par\medskip\raggedright\noindent\normalsize\acceptededitorial{Table~\ref{tab:distribution-details} evaluates human distributional fidelity and coverage. AESOP improves all four PRDC measures, with recall of 0.835 versus 0.215/0.312 for DiT/MAR. Higher precision accompanies this broader coverage, complementing the lower human distribution distance in Table~\ref{tab:main}.}\par
\end{table}

\subsection{\acceptededitorial{Representation and camera control}}

\paragraph{Representation quality.}
\acceptededitorial{The AE rows in Table~\ref{tab:main} measure reconstruction by encoding and decoding reference sequences. AESOP reduces camera-center ADE to 0.029 m and rotation error to $0.41^\circ$, compared with 0.110 m and $1.37^\circ$ for PulpMotion. It also yields lower human and camera distribution distances, whereas PulpMotion has higher human-text alignment (20.15 versus 17.56). Supplementary Table~\ref{tab:ae-reconstruction} reports additional human positional and camera-motion errors.}

\begin{figure}[!t]
\centering
\includegraphics[width=\linewidth]{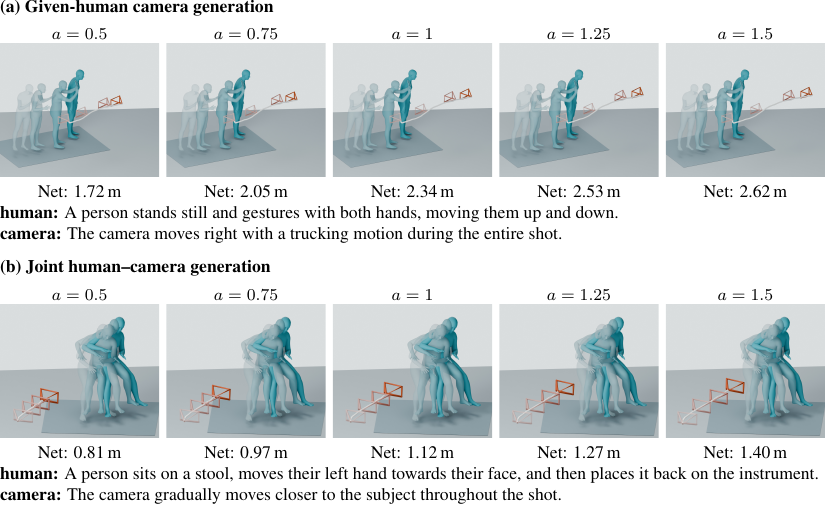}\par\vspace{-1.84pt}
\caption{\acceptededitorial{\textbf{Translation-intensity examples} with shared human context, text and camera noise. Outputs are regenerated at each intensity; numbers report first-to-last camera-center displacement. Fading snapshots indicate temporal order. Global views are fitted separately to each trajectory\textquotesingle s spatial extent.}}
\label{fig:intensity-examples}
\end{figure}

\paragraph{Translation intensity.}
\acceptededitorial{Figures~\ref{fig:intensity-examples} and~\ref{fig:population-intensity} show qualitative examples and an eleven-value scan from $a=0.5$ to $1.5$ (step 0.1). Path length sums camera-center travel; net displacement measures endpoint distance. Mean path length rises from 0.647 to 1.023 m (given-human) and from 0.367 to 0.612 m (joint). We require travel to be nondecreasing in all ten adjacent intervals: 81.50\%/81.37\% of given-human/joint clips satisfy this for path length, and 78.61\%/78.46\% for net displacement. For the 2,106 clips with reference net displacement $\geq0.2$ m, path-length rates reach 99.19\%/98.86\%. Supplementary Section~\ref{supp:intensity} reports endpoint results.}

\acceptededitorial{CLaTr and F1 increase alongside camera travel, while higher r-FPD and Out indicate a framing trade-off. Given-human ADE and rotation error in both tasks vary little. Given-human FDC is lowest near $a=0.8$, whereas joint FDC decreases across the grid.}

\begin{figure}[!htbp]
\centering
\includegraphics[width=\textwidth]{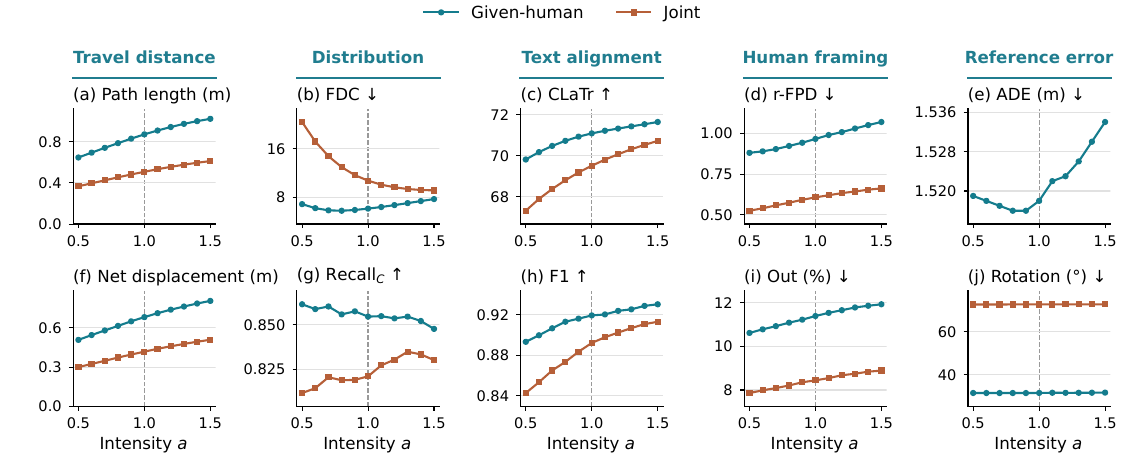}
\caption{\acceptededitorial{\textbf{Translation-intensity response and camera quality.} Columns group travel distance, distribution, text alignment, framing and reference error. Results use one training seed and shared human context, text and camera noise across intensities. ADE is reported for given-human generation. Vertical dashed lines mark $a=1$.}}
\label{fig:population-intensity}
\label{fig:intensity-quality}
\end{figure}

\begin{samepage}
\paragraph{Geometry-grounded training.}
\acceptededitorial{We evaluate direction accuracy on original and direction-reversed prompts, comparing translation signs over reference-active axes and time steps. After DPA continuation, AESOP reaches 84.33\%/83.11\% combined accuracy in given-human/joint generation, exceeding the matched factual-pair control by 1.67/1.38 percentage points (Supplementary Table~\ref{tab:dpa-direction}). Accuracy improves in both prompt arms: on reversed prompts, DPA scores 79.00\%/79.58\%, compared with 77.71\%/78.51\% for factual continuation. IPA then adds continuous intensity control, with similar default camera quality to the +DPA endpoint in Table~\ref{tab:main}.}

\par\end{samepage}

\subsection{\acceptededitorial{User study}}
\label{sec:user-study}
\begin{table}[H]\begin{minipage}[t]{0.49\linewidth}\vspace{0pt}
\setlength{\abovecaptionskip}{0pt}\setlength{\belowcaptionskip}{6pt}
\caption{\acceptededitorial{Overall-quality preference for AESOP (\%). Scores average clip-level responses, with ties receiving half credit. Intervals are participant-cluster 95\% CIs.}}
\label{tab:user-study-overall}

\centering\begingroup\fontsize{7}{8.4}\selectfont
\setlength{\tabcolsep}{2pt}\renewcommand{\arraystretch}{1.2}
\begin{tabular*}{\linewidth}{@{\extracolsep{\fill}}lrr@{}}\toprule
Compared with & Pref. & 95\% CI\\\midrule
\multicolumn{3}{l}{\textit{Given-human}}\\
CCD & 75.3 & [64.8, 85.5]\\
DanceCamera3D & 99.2 & [97.5, 100.0]\\
DIRECTOR-C & 86.6 & [81.5, 91.5]\\
\midrule\multicolumn{3}{l}{\textit{Joint}}\\
PulpMotion DiT & 86.3 & [80.1, 91.9]\\
PulpMotion MAR & 92.3 & [86.9, 96.9]\\
\bottomrule\end{tabular*}\endgroup

\end{minipage}\hfill
\begin{minipage}[t]{0.49\linewidth}\vspace{0pt}
\acceptededitorial{We conducted an anonymized paired-comparison study with 29 participants. Each participant completed 20 given-human and 18 joint comparisons, with balanced method pairs and randomized presentation. Participants rate overall camera quality for given-human generation and overall human--camera quality for joint generation. Preference, averaged over clips and equally over opponents, is 87.0\% (95\% CI: 83.8--90.8) and 89.3\% (84.4--93.6), respectively. Table~\ref{tab:user-study-overall} gives the five pairwise comparisons; Supplementary Section~\ref{supp:user-study} details the protocol and individual criteria.}
\end{minipage}
\end{table}

\par\noindent\begin{minipage}{\textwidth}\section{Conclusion}
\label{sec:conclusion}

\acceptededitorial{AESOP unifies given-human camera generation and joint human--camera generation through an independent human pathway and a shared human-conditioned camera generator. Geometry-grounded intensity pairs teach explicit translation-intensity control, complemented by auxiliary direction supervision. Experiments on PulpMotion show strong camera distributional and framing quality in both tasks, while the learned intensity condition adjusts camera travel.}

\paragraph{Limitations and future directions.}
\label{sec:limitations}
The sequence-level intensity condition limits control over individual events, while complete human context and offline sampling limit streaming generation. Future directions include event-wise intensity control and causal, incremental camera generation.
\end{minipage}\par

\FloatBarrier
\endgroup
\bibliographystyle{iclr2027_conference}
\bibliography{storymotion_references}

\clearpage
\appendix
\setcounter{table}{0}
\setcounter{figure}{0}
\setcounter{equation}{0}
\renewcommand{\thetable}{S\arabic{table}}
\renewcommand{\thefigure}{S\arabic{figure}}
\renewcommand{\theequation}{S\arabic{equation}}

\renewcommand{\theHtable}{supp.\arabic{table}}
\renewcommand{\theHfigure}{supp.\arabic{figure}}
\renewcommand{\theHequation}{supp.\arabic{equation}}
\newcommand{\sg}{\operatorname{sg}}
\newcommand{\mmean}{\operatorname{MaskedMean}}
\newcommand{\sml}{\operatorname{MaskedSmoothL1}}
\newcommand{\mse}{\operatorname{MaskedMSE}}
\raggedbottom

\setcounter{topnumber}{6}
\setcounter{bottomnumber}{6}
\setcounter{totalnumber}{10}
\renewcommand{\topfraction}{0.95}
\renewcommand{\bottomfraction}{0.95}
\renewcommand{\textfraction}{0.05}
\renewcommand{\floatpagefraction}{0.8}
\setlength{\textfloatsep}{10pt plus 2pt minus 2pt}
\setlength{\intextsep}{10pt plus 2pt minus 2pt}
\setlength{\floatsep}{10pt plus 2pt minus 2pt}
\makeatletter
\setlength{\@fptop}{0pt}
\setlength{\@fpsep}{12pt}
\setlength{\@fpbot}{0pt plus 1fil}
\makeatother
\etocdepthtag{supplement}
\etocsettagdepth{main}{none}
\etocsettagdepth{supplement}{subsection}
\renewcommand{\contentsname}{Supplementary Contents}
\tableofcontents
\clearpage

\section{Disclosure of AI Use}
\label{supp:ai-disclosure}
The authors conceived the research ideas, designed the methods, implemented the core code, and analyzed the experimental results. Under the authors' direction, generative AI tools assisted with supporting code implementation, experimental execution, threshold selection for DPA/IPA data construction, and language refinement. The authors made the final research decisions and take full responsibility for the methods, implementation, experimental results, analyses, and final manuscript.

\section{\acceptededitorial{Representation and Implementation}}
\label{supp:representation}\label{supp:losses}
\subsection{Symmetric and asymmetric representations}
\label{supp:interface}
\paragraph{PulpMotion's symmetric representation.}

PulpMotion's shared encoder~\citep{courant2026pulpmotion} first maps human motion and camera trajectory to a joint latent, which is then split into human and camera components:
\begin{equation}
 z_{HC}^{P}=E_{HC}(H,C),\quad (z_H^{P},z_C^{P})=\operatorname{split}(z_{HC}^{P}),\quad z_F^{P}=Wz_{HC}^{P}.
\end{equation}
Independent human and camera decoders reconstruct their respective inputs from $z_H^{P}$ and $z_C^{P}$. A third decoder reconstructs framing from the auxiliary latent $z_F^{P}$, obtained by a learned linear map $W$. The shared encoder and all three reconstruction branches are trained together. Both latents can depend on both inputs: camera reconstruction has access to human context, while human reconstruction also depends on camera input through the shared encoder. Consequently, human representation learning is coupled to camera training, and human encoding requires camera input.

\begin{figure}[!htbp]
\centering
\includegraphics[width=\linewidth]{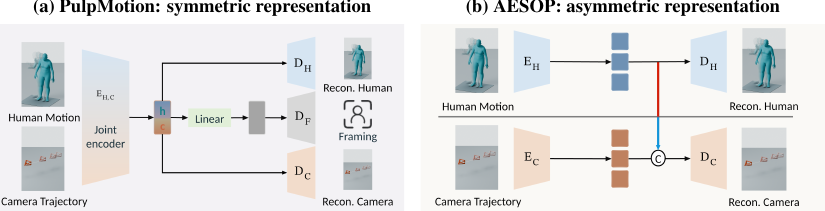}
\caption{\textbf{human--camera representation dependencies.} (a) PulpMotion's symmetric representation encodes both inputs with a shared encoder, reconstructs each from its latent, and learns an auxiliary framing latent from their concatenation~\citep{courant2026pulpmotion}. (b) AESOP separately encodes the two inputs. The human decoder reads only human latents; the camera decoder concatenates camera latents with frozen human latents (circled C). Human representation training is completed before camera training.}
\label{fig:representation-comparison}
\end{figure}

\begin{table}[!htbp]
\setlength{\abovecaptionskip}{0pt}\setlength{\belowcaptionskip}{6pt}

\makeatletter
\long\def\@makecaption#1#2{
  \vskip\abovecaptionskip
  \noindent #1: #2\par
  \vskip\belowcaptionskip}
\makeatother
\caption{\textbf{Training and reconstruction dependencies.} Both representations provide human context for camera reconstruction. AESOP additionally separates human learning from camera training and makes human reconstruction independent of camera input.}
\label{tab:representation-design}

\centering\storytablefont
\setlength{\tabcolsep}{4pt}\renewcommand{\arraystretch}{1.2}
\begin{tabularx}{\textwidth}{@{}l>{\hsize=1.2\hsize\linewidth=\hsize\raggedright\arraybackslash}X*{2}{>{\hsize=.9\hsize\linewidth=\hsize\centering\arraybackslash}X}@{}}
\toprule
 & Training & \multicolumn{2}{c}{Reconstruction}\\
\cmidrule(lr){2-2}\cmidrule(lr){3-4}
Method & Independent human learning & camera-independent human & human-aware camera\\\midrule
PulpMotion (symmetric) & \storyno\ Both streams trained together & \storyno & \storyyes\\
\rowcolor{storymotionblue}
AESOP (asymmetric) & \storyyes\ Train human first, then freeze & \storyyes & \storyyes\\
\bottomrule
\end{tabularx}

\end{table}

\paragraph{AESOP's asymmetric representation.}
\acceptededitorial{AESOP retains PulpMotion's human and camera feature format but uses separate encoders (Fig.~\ref{fig:representation-comparison}). Its human decoder reads only the human latent; the camera decoder reads the concatenated human and camera latents.}

\begin{table}[!t]
\setlength{\abovecaptionskip}{0pt}\setlength{\belowcaptionskip}{6pt}

\makeatletter
\long\def\@makecaption#1#2{
  \vskip\abovecaptionskip
  \noindent #1: #2\par
  \vskip\belowcaptionskip}
\makeatother
\caption{\acceptededitorial{\textbf{Autoencoder reconstruction on the test split.} Sym./asym. denote PulpMotion\textquotesingle s symmetric and AESOP\textquotesingle s asymmetric autoencoders. Motion errors are clip-balanced and measured in mm/s, $^\circ$/s and m/s$^3$, respectively. MPJPE, ADE and FDE are in meters; aligned MPJPE removes root translation. Bold marks the best value between the autoencoders.}}
\label{tab:ae-reconstruction}

\centering\begingroup\fontsize{7}{8.4}\selectfont
\setlength{\tabcolsep}{1pt}\renewcommand{\arraystretch}{1.2}
\begin{tabularx}{\textwidth}{@{}l*{8}{>{\raggedleft\arraybackslash}X}@{}}
\toprule
 & \multicolumn{4}{c}{human reconstruction} & \multicolumn{4}{c}{camera reconstruction}\\
\cmidrule(lr){2-5}\cmidrule(lr){6-9}
Autoencoder & Global MPJPE$\downarrow$ & Aligned MPJPE$\downarrow$ & Root ADE$\downarrow$ & Root FDE$\downarrow$ & FDE$\downarrow$ & Trans. vel.$\downarrow$ & Rot. vel.$\downarrow$ & Trans. jerk$\downarrow$\\\midrule
PulpMotion (sym.) & \metricstd{0.1620}{0.02} & \metricstd{0.0617}{0.01} & \metricstd{0.1376}{0.01} & \metricstd{0.3283}{0.04} & \metricstd{0.1677}{0.02} & \metricstd{124.28}{3.02} & \textbf{\metricstd{10.55}{0.10}} & \metricstd{154.67}{1.54}\\
\rowcolor{storymotionblue}
AESOP (asym.) & \textbf{\metricstd{0.1304}{0.01}} & \textbf{\metricstd{0.0479}{0.003}} & \textbf{\metricstd{0.1063}{0.01}} & \textbf{\metricstd{0.2611}{0.02}} & \textbf{\metricstd{0.0356}{0.002}} & \textbf{\metricstd{10.11}{0.81}} & \metricstd{10.98}{0.21} & \textbf{\metricstd{8.55}{0.40}}\\
\bottomrule
\end{tabularx}\endgroup

\end{table}

\acceptededitorial{Table~\ref{tab:ae-reconstruction} evaluates deterministic encoding and decoding. AESOP has lower human positional, translation-velocity and jerk reconstruction errors, while PulpMotion has lower rotation-velocity error. Root-aligned human error is smaller than global error for both representations, indicating the contribution of root-trajectory deviations. Table~\ref{tab:main} reports distributional and camera-pose reconstruction quality.}

\subsection{Implementation, training objectives and inference}
\label{supp:training}
\paragraph{Per-frame features.}
We use PulpMotion's per-frame feature format~\citep{courant2026pulpmotion}, whose human features follow the SMPL-based representation of \citet{add_petrovich2024stmc}; SMPL is defined by \citet{add_loper2015smpl}. Each human frame has 199 features describing root height, planar root velocity, yaw velocity, local joint rotations and local joint positions. Each camera frame has 14 features: two field-of-view angles, three camera-minus-human relative-position channels, a six-dimensional rotation representation~\citep{smadd_zhou2019rotation} and three world-space translation increments.

\paragraph{Latent interface.}
For a sequence of $T$ frames, the human and camera encoders produce $\lceil T/4\rceil$ latent tokens with 128 and 64 channels, respectively. The human decoder reads only the 128-channel human latent; the camera decoder reads the 192-channel concatenation of human and camera latents.

Noncausal width-256 temporal encoders use stride four; transposed-convolution decoders crop to the requested length. A valid-frame mask marks actual sequence frames and excludes batch padding from losses; latent masks retain $\lceil T/4\rceil$ tokens. Both flow Transformers have 12 layers, width 512, eight attention heads, FFN multiplier four and dropout 0.1. Camera blocks apply self-attention, camera-text attention, full-human attention and FFN in order, using frozen 512-dimensional CLIP text features~\citep{smadd_radford2021clip}.

\acceptededitorial{For stream $k\in\{H,C\}$, let $\widetilde z_k$ denote the encoder output and $z_k$ the normalized latent used by the flow. Normalization uses training-only channel statistics $\mu_k,s_k$ and standardized-stream mean/covariance factors $m_k,L_k$. Cholesky whitening uses ridge $10^{-4}$; its inverse precedes decoding:
\begin{equation}
 z_k=L_k^{-1}\bigl((\widetilde z_k-\mu_k)/s_k-m_k\bigr),\qquad
 \widetilde z_k=\mu_k+s_k\odot(m_k+L_kz_k).
\end{equation}}
Invalid positions are zeroed after either operation. Physical decoding integrates de-normalized translation increments and anchors the first camera center using the decoded relative vector and human root~\citep{courant2026pulpmotion}, using predicted features for initialization. Human-feature, relative-center and translation-increment normalization are distinct from latent whitening.

\paragraph{Stage 1 objectives.}
Let $M$ mark valid frames and $M_{\Delta,t}=M_tM_{t-1}$ for $t=1,\ldots,T-1$. Angle brackets average selected scalar elements. For normalized features $H,C$, cumulative de-normalized yaw $\psi$, and decoded root trajectory $r$, the complete reconstruction losses are
\begin{equation}
\begin{aligned}
\mathcal L_H^{\mathrm{rec}}={}&\langle\ell_{\mathrm{SL1}}(\hat H,H)\rangle_M
 +\langle(\Delta\hat H-\Delta H)^2\rangle_{M_\Delta}\\
 &+10^{-3}\langle1-\cos(\hat\psi-\psi)\rangle_M
 +3\!\times\!10^{-3}\langle\ell_{\mathrm{SL1}}(\hat r,r)\rangle_M,\\
\mathcal L_C^{\mathrm{rec}}={}&\langle\ell_{\mathrm{SL1}}(\hat C,C)\rangle_M
 +\langle(\Delta\hat C-\Delta C)^2\rangle_{M_\Delta}.
\end{aligned}
\end{equation}

\paragraph{Stage 2 objectives and optimization.}
We sample $u\sim\mathcal U(0,1)$, set $\sigma=5u/(1+4u)$ and draw $\epsilon\sim\mathcal N(0,I)$. \acceptededitorial{The velocity target $v^*$ follows Eq.~\ref{eq:human-camera-flow-matching}.} For $n_b$ valid latent tokens and stream dimension $D$, define
\begin{equation}
 e_b=\frac{\sum_{d,t}M_{b,t}(v_{b,d,t}-v^*_{b,d,t})^2}{Dn_b},\quad
 \mathcal L_{\mathrm{original}}=\frac1B\sum_b e_b,\quad
 \mathcal L_{\mathrm{continuation}}=\frac{\sum_b n_b e_b}{\sum_b n_b}.
\end{equation}
Continuation weights clips by length; original training weights them equally.
Each continuation batch has 120 rows, with two rows per control pair. During the 35K DPA phase, direction-pair slots rise from zero to five over the first 2K updates, remain at five through 20K, fall to two through 30K and then to one. During the 15K IPA phase, each batch retains two DPA pair slots; active IPA slots rise to four and null IPA slots to one during the first 1K updates.

\begin{table}[ht]
\setlength{\abovecaptionskip}{0pt}\setlength{\belowcaptionskip}{6pt}

\makeatletter
\long\def\@makecaption#1#2{
  \vskip\abovecaptionskip
  \noindent #1: #2\par
  \vskip\belowcaptionskip}
\makeatother
\caption{\acceptededitorial{Training schedule. The final phase additionally trains the intensity MLP at $10^{-4}$.}}
\label{tab:optimization}

\centering\storytablefont\setlength{\tabcolsep}{3pt}
\begin{tabular*}{\linewidth}{@{\extracolsep{\fill}}llrrll@{}}
\toprule
Stage & Phase & Updates & Batch & Learning rate & Frozen components\\\midrule
Reconstruction & human autoencoder & 210K & 128 & $5\times10^{-5}$ & None\\
 & camera autoencoder & 210K & 128 & $5\times10^{-5}$ & human autoencoder\\\midrule
Generation & human flow & 105K & 128 & $2\times10^{-4}$ & Autoencoders\\
 & Original camera & 105K & 128 & $10^{-4}$ & Autoencoders, human flow\\
 & camera + DPA & 35K & 120 & $2\times10^{-5}$ & Autoencoders, human flow\\
 & camera + DPA + IPA & 15K & 120 & $2\times10^{-5}$ & Autoencoders, human flow\\\bottomrule
\end{tabular*}

\end{table}
Stage~1 uses AdamW~\citep{add_loshchilov2019adamw} with $\beta=(0.9,0.999)$, zero weight decay, a 1K-step linear warmup and cosine decay to $10^{-6}$ per phase. Stage~2 uses AdamW with $\beta=(0.9,0.95)$, weight decay 0.01, gradient clipping 1.0 and EMA decay 0.9999. The human flow warms up for 2K steps and decays its rate by 0.1 at step 80K; original camera training uses a constant rate. Continuations use BF16 and 1K-step warmup.

\paragraph{Inference and conditioning.}
\acceptededitorial{Sampling follows Section~\ref{sec:setup} with solver shift five and camera-text guidance from Eq.~\ref{eq:camera-cfg}. The intensity MLP described in Section~\ref{sec:method} has 64 hidden units and a 512-dimensional output added to the camera timestep condition.}

\section{DPA and IPA Data Construction}
\label{supp:data}
\begin{figure}[!htbp]
\centering
\includegraphics[width=\textwidth]{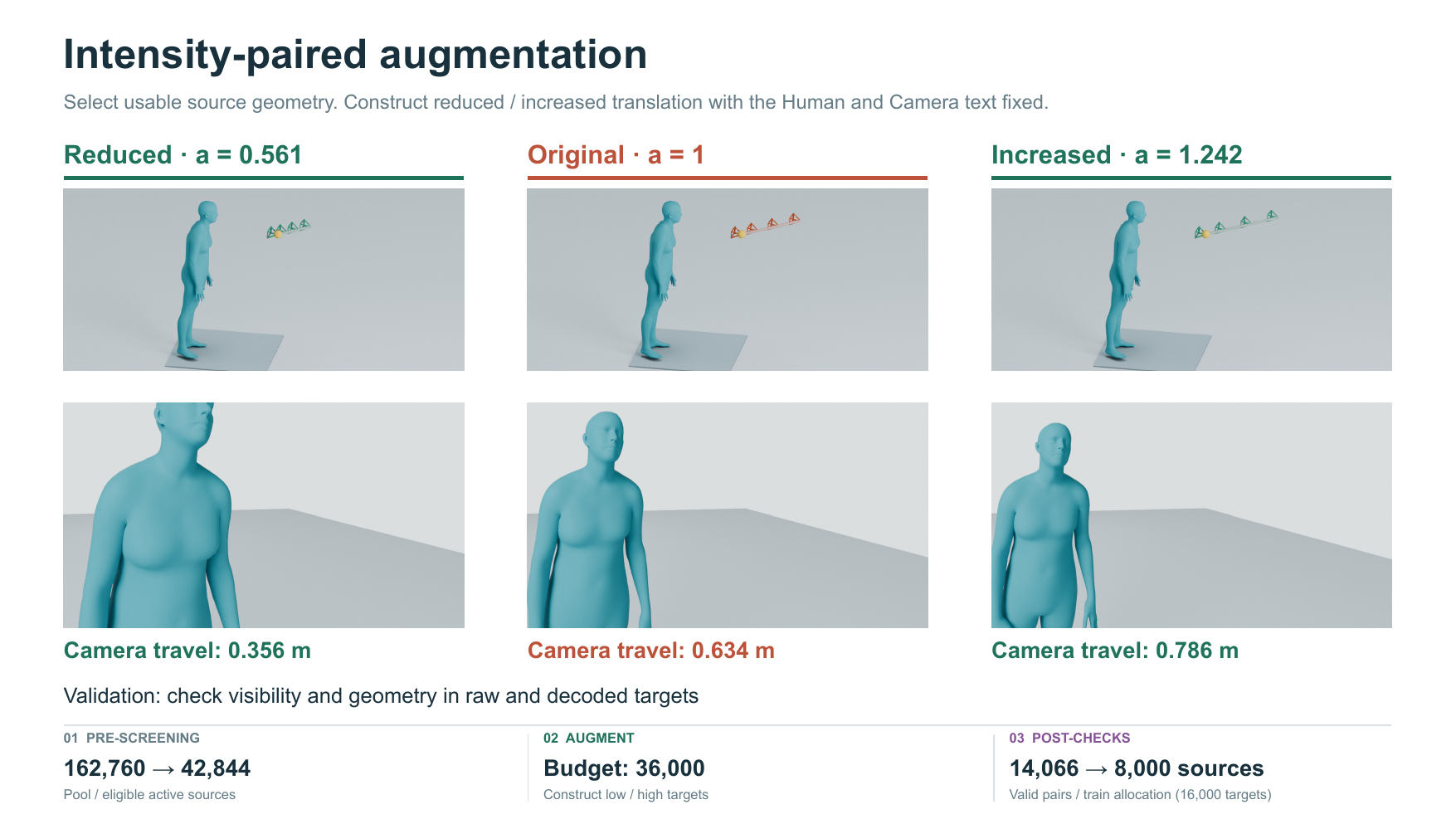}
\caption{\textbf{IPA target construction.} Reduced, original and increased translation share human motion and camera text. Camera travel changes from 0.356 to 0.634 to 0.786 m in this example. The gold point marks the initial camera center; global views and synchronized final-frame projections show the resulting geometry and framing. The three stages summarize the active branch, which supplies 16,000 training targets from 8,000 sources. The accompanying null branch supplies 2,000 low-translation sources with unchanged targets.}
\label{fig:cts-filtering}
\end{figure}

We construct paired camera targets from the 162,760-clip PulpMotion training pool. DPA changes translation direction and its text condition; IPA changes translation intensity while keeping camera text fixed. \acceptededitorial{Both preserve human motion, camera rotation, field of view and timing; the human latent and decoded output are unchanged across each constructed pair.} Figures~\ref{fig:cf-filtering} and~\ref{fig:cts-filtering} illustrate the three construction stages and the retained source counts. Table~\ref{tab:data-gates} lists the acceptance rules.

\subsection{Direction-paired augmentation}
\label{supp:dpa-construction}
\paragraph{1. Pre-screening.}
From the training pool, DPA selects 22,783 whole-shot captions containing one Truck or Dolly event. Source-motion checks retain 9,143 clips with stable, sufficient translation. Truck uses a fixed camera-local right axis, obtained by averaging the first $\min(8,\lfloor T/5\rfloor)$ right vectors, projecting onto the ground plane and normalizing. Dolly uses camera--human root distance.

\paragraph{2. Augmentation.}
On the accepted interval $[t_s,t_e]$, we reflect the signed coordinate as $q'_t=2q_{t_s}-q_t$. Truck preserves orthogonal components; Dolly preserves the human-to-camera unit direction. The trajectory is unchanged before the event and carries its endpoint offset afterward. Relative-center and velocity features are recomputed. Paired minimal text templates differ only in direction.

\begin{figure}[!htbp]
\centering
\includegraphics[width=\textwidth]{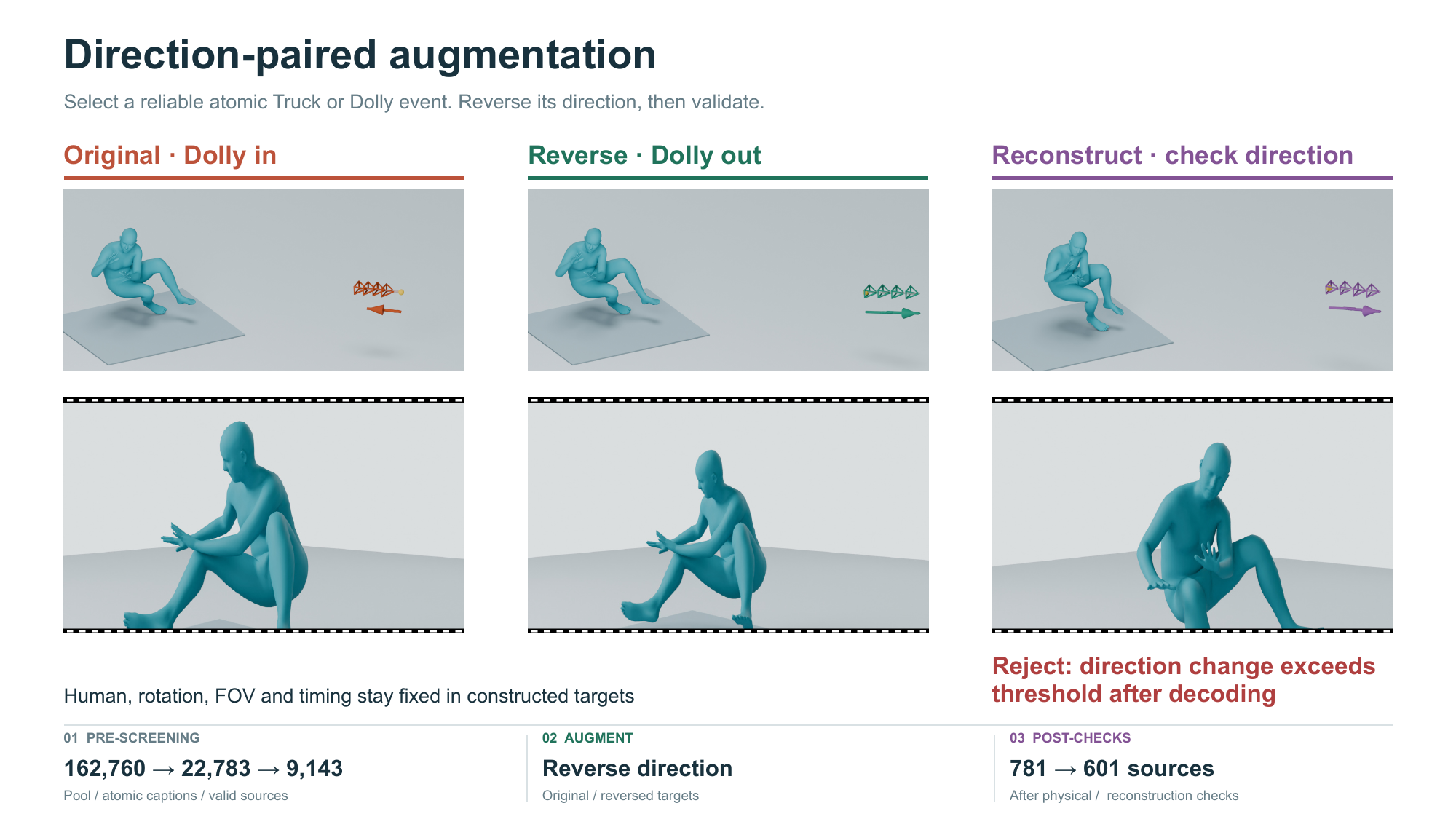}
\caption{\textbf{DPA construction and validation.} Left: the original Dolly-in trajectory. Middle: its opposite Dolly-out target, with human motion, camera rotation, field of view and timing fixed. Right: a reconstruction rejected because its direction change exceeds the acceptance threshold after decoding. The gold point marks the initial camera center; global views show the starting human pose and full trajectory, and projections show the synchronized final frame. The three stages summarize source filtering, direction reversal and target checks; 601 sources pass the final reconstruction check.}
\label{fig:cf-filtering}
\end{figure}

\paragraph{3. Post-checks.}
Physical checks assess geometry, framing, visible direction effect and non-target leakage, retaining 781 constructed sources (8.54\% of the 9,143 eligible sources). Reconstruction checks assess fidelity, direction, response and visibility, yielding 601 sources: 375 Dolly and 226 Truck. These supply 601 original/opposite training pairs with 1,202 targets. Figure~\ref{fig:cf-filtering} illustrates why the decoded direction is checked as well as the raw target.

\acceptededitorial{The DPA visible-effect score is $V=\max(\Delta x/0.08,\Delta\alpha/6^\circ)$ for Truck and $V=\Delta h/0.12$ for Dolly. Here $\Delta x$ is peak projected subject-center separation in image-width units, $\Delta\alpha$ is peak human-relative azimuth separation, and $\Delta h=\max_t|h'_t/h_t-1|$ is peak projected box-height ratio change, comparing original/opposite targets within the event. Decoded retention compares this score after versus before reconstruction.}

\FloatBarrier
\subsection{Intensity-paired augmentation}
\label{supp:ipa-construction}
\paragraph{1. Pre-screening.}
The active branch selects trajectories with sufficient translation to provide weaker/stronger targets. The null branch selects low-translation trajectories and pairs an unchanged target with different intensities, teaching the model to retain low-translation shots under intensity changes. Motion, visibility and distance checks, together with DPA-source exclusion, yield 42,844 active and 75,863 null candidates. Deterministic hash ranking selects 36,000 active sources for target construction.

\paragraph{2. Augmentation.}
\acceptededitorial{An 11-frame triangular filter with reflection padding decomposes camera centers into $p_t=s_t+r_t$. We apply Eq.~\ref{eq:geometry-programs} and recompute relative-center and velocity features from the transformed trajectory.}

\acceptededitorial{For IPA target construction, candidate labels are sampled from $0<a<1$ for weaker targets and $1<a<2$ for stronger targets.} Each interval is divided into four equal-width bins, ordered from smaller to larger $a$. Their sampling probabilities are $(0.1,0.2,0.3,0.4)$ below one and $(0.4,0.3,0.2,0.1)$ above one, favoring labels closer to $a=1$. A fixed-seed sampler selects a bin and then a continuous value within it. For each active source, we try up to eight labels per interval and keep the first weaker and first stronger target that pass the checks below. Null sources retain the unchanged camera target while receiving alternating weaker and stronger labels from the same distributions. The eleven-level evaluation uses the narrower range $a\in[0.5,1.5]$, in steps of $0.1$.

\paragraph{3. Post-checks.}
We retain active sources with an accepted target on both sides of $a=1$. Physical and construction-representation reconstruction checks yield 14,066 sources (39.07\% of the 36,000 constructed sources). After reserving source-video groups for development, the final training set contains 8,000 active sources, yielding 16,000 camera targets labeled $a\ne1$, and 2,000 null sources. Development control variants are kept separate from control training and \acceptededitorial{the test split}.

\begin{table}[!htbp]
\setlength{\abovecaptionskip}{0pt}\setlength{\belowcaptionskip}{6pt}

\makeatletter
\long\def\@makecaption#1#2{
  \vskip\abovecaptionskip
  \noindent #1: #2\par
  \vskip\belowcaptionskip}
\makeatother
\caption{\acceptededitorial{Construction acceptance criteria for DPA and IPA, grouped by source selection, target geometry, framing and reconstruction.}}
\label{tab:data-gates}

\centering\begingroup\storytablefont\renewcommand{\arraystretch}{1.20}
\setlength{\tabcolsep}{4pt}
\begin{tabular*}{\linewidth}{@{\extracolsep{\fill}}>{\raggedright\arraybackslash}p{.21\linewidth}>{\raggedright\arraybackslash}p{.75\linewidth}@{}}
\toprule
Operation & Acceptance rules\\\midrule
\multicolumn{2}{l}{\textit{DPA: direction pairs}}\\\midrule
Select source events & Duration $\ge45$ frames; same-sign velocity fraction $\ge0.80$; signed net displacement/path $\ge0.60$. Truck displacement $\ge0.20$ m; Dolly displacement $\ge\max(0.20\text{ m},0.10\rho_{t_s})$.\\[4pt]
Check target geometry & Opposite-target human distance $>0.25$ m; opposite/original response $\in[0.80,1.25]$. Rotation/FOV unchanged; Truck vertical shift $\le10^{-5}$ m and radial relative change $\le0.10$; Dolly radial-direction change $\le0.05^\circ$.\\[4pt]
Check framing and effect & No zero-visible frame; visibility $\ge0.80$; opposite outscreen increase $\le0.05$. Truck: projected center shift $\ge0.08$ image width or viewpoint change $\ge6^\circ$. Dolly: peak box-height ratio change $\ge0.12$.\\[4pt]
Check reconstructed targets & Feature MSE $\le\acceptededitorial{0.01181}$; both direction signs correct; decoded/raw response ratios $\in[0.70,1.30]$. No framing failure; $V_{\rm dec}\ge1$ and $V_{\rm dec}/V_{\rm raw}\ge0.70$.\\\midrule
\multicolumn{2}{l}{\textit{IPA: intensity pairs}}\\\midrule
Select sources & Duration $\ge1$ s. Active: path $\ge\acceptededitorial{0.4118}$ m, span $\ge0.15$ m, max-step/path $\le0.20$. Null: path $\le0.10$ m. Source visibility $\ge0.80$ and human distance $\ge0.25$ m; exclude DPA sources.\\[4pt]
Check magnitude and direction & Active only: absolute requested/actual path-ratio error $\le0.08$; path/speed-ratio difference $\le0.03$; direction cosine $\ge0.90$ for $a\ge0.25$; $|a-1|\times$ original path $\ge0.15$ m.\\[4pt]
Check dynamics and framing & Active only: speed/acceleration/jerk p95 $\le\acceptededitorial{0.1416}$ m/frame, $\acceptededitorial{0.03499}$ m/frame$^2$, $\acceptededitorial{0.01315}$ m/frame$^3$. Human distance $\in[0.25,\acceptededitorial{12.7716}]$ m; visibility decrease $\le0.05$; severe-outscreen increase $\le0.03$.\\[4pt]
Check reconstruction & Active: construction-representation decoded path-ratio error $\le0.15$.\\
\bottomrule
\end{tabular*}\endgroup

\end{table}

\FloatBarrier
\begin{samepage}
\paragraph{Construction checks.}
\acceptededitorial{Table~\ref{tab:data-gates} groups the acceptance criteria. Dynamics limits use the 99.5th percentiles of a 1,000-clip training-data pilot, in per-frame units. A joint is visible when its projection is finite, lies inside the image and has camera-space depth $>10^{-4}$ m. DPA visibility averages this indicator over joints and frames; outscreen is its complement. IPA visibility counts frames with any visible joint; severe-outscreen counts frames with fewer than half the joints visible. The decoded IPA gate uses the construction representation, and null targets remain unchanged.}

\par\end{samepage}
\section{Evaluation and Baseline Protocols}
\label{supp:evaluation}\label{supp:baselines}
\subsection{\acceptededitorial{Baseline protocols}}
\label{supp:baseline-protocols}
\acceptededitorial{We follow the evaluation setup in Section~\ref{sec:setup} and the AESOP training schedule in Table~\ref{tab:optimization}. Given-human evaluation supplies reference human motion; joint evaluation uses each method\textquotesingle s generated motion. Table~\ref{tab:baseline-protocol} lists adaptations and sampling configurations.}

\begin{table}[H]
\setlength{\abovecaptionskip}{0pt}\setlength{\belowcaptionskip}{6pt}

\makeatletter
\long\def\@makecaption#1#2{
  \vskip\abovecaptionskip
  \noindent #1: #2\par
  \vskip\belowcaptionskip}
\makeatother
\caption{\acceptededitorial{\acceptededitorial{Baseline inputs, conditioning and sampling. \acceptededitorial{For PulpMotion, the tuple denotes conditional-text guidance, autoregressive-context guidance, and auxiliary projection guidance, respectively. A negative autoregressive-context weight selects standard CFG; zero auxiliary projection weight disables that guidance term.} MAR generates autoregressively, with a diffusion sampler inside each iteration. EDM, DDIM and DDPM follow \citet{add_karras2022edm}, \citet{add_song2021ddim} and \citet{add_ho2020ddpm}, respectively.}}}
\label{tab:baseline-protocol}

\centering\storytablefont\setlength{\tabcolsep}{3pt}
\begin{tabular*}{\linewidth}{@{\extracolsep{\fill}}>{\raggedright\arraybackslash}p{.20\linewidth}>{\raggedright\arraybackslash}p{.36\linewidth}>{\raggedright\arraybackslash}p{.37\linewidth}@{}}
\toprule
Method & Inputs and conditioning & Sampling and guidance\\\midrule
DIRECTOR-C & \acceptededitorial{Camera text and human root trajectory} & \acceptededitorial{EDM: 10 Euler/Heun steps; camera CFG 1.4}\\
DanceCamera3D & \acceptededitorial{Camera text and full human motion; text replaces music} & \acceptededitorial{DDIM: 50 steps, $\eta=1$; human/camera CFG 1.75/1}\\
CCD & \acceptededitorial{Camera text; text-only network} & \acceptededitorial{DDPM: 1,000 steps; camera CFG 2}\\
CCD-H & \acceptededitorial{Camera text and human joints; text cross-attention} & \acceptededitorial{DDPM: 1,000 steps; camera CFG 2}\\
PulpMotion DiT & \acceptededitorial{Human and camera text; symmetric representation} & \acceptededitorial{DDPM: 50 steps; \acceptededitorial{$(g_c,g_m,g_z)=(11,-1,0)$}}\\
PulpMotion MAR & \acceptededitorial{Human and camera text; symmetric representation} & \acceptededitorial{18 autoregressive iterations, each with 50 DDPM steps; \acceptededitorial{$(g_c,g_m,g_z)=(3.5,2,0)$}}\\
\rowcolor{storymotionblue}
AESOP & \acceptededitorial{Human motion and camera text (given); human text and camera text (joint)} & \acceptededitorial{Euler: 50 steps per stream; human/camera CFG 1/1.5}\\\bottomrule
\end{tabular*}

\end{table}

\begin{samepage}
\acceptededitorial{Generative training presents 13.44M examples to each camera-only baseline and 26.88M to each PulpMotion joint model. AESOP presents 13.44M examples to each human/camera flow, followed by 6M camera continuation examples. These counts include repeated presentations and both members of control pairs; Table~\ref{tab:optimization} gives AESOP\textquotesingle s phase durations and batch sizes.}
\par\end{samepage}

\paragraph{\acceptededitorial{Human-relative camera decoding.}}
\acceptededitorial{CCD and CCD-H decode camera position relative to the supplied human head in a heading-aligned frame, with a head-directed look-at rotation adjusted by predicted screen position~\citep{jiang2024ccd}. AESOP and PulpMotion anchor camera position to the human root at the first frame and accumulate world-space increments~\citep{courant2026pulpmotion}.}

\subsection{Direction-paired training}
\label{supp:dpa-evaluation}
\begin{table}[!htbp]
\setlength{\abovecaptionskip}{0pt}\setlength{\belowcaptionskip}{6pt}

\makeatletter
\long\def\@makecaption#1#2{
  \vskip\abovecaptionskip
  \noindent #1: #2\par
  \vskip\belowcaptionskip}
\makeatother
\caption{\acceptededitorial{\acceptededitorial{Direction accuracy (\%).} DPA and factual continuations share the starting checkpoint, 35K updates, batch allocation and sampling schedule; factual pairs replace directional counterfactuals in the control. Bold marks the best mean in each column. Section~\ref{supp:dpa-evaluation} defines paired sampling and scoring.}}
\label{tab:dpa-direction}

\centering\begingroup\storytablefont
\setlength{\tabcolsep}{3pt}\renewcommand{\arraystretch}{1.15}
\begin{tabular*}{\textwidth}{@{\extracolsep{\fill}}lrrrrrr@{}}
\toprule
 & \multicolumn{3}{c}{Given-human generation} & \multicolumn{3}{c}{Joint generation}\\
\cmidrule(lr){2-4}\cmidrule(lr){5-7}
Checkpoint & Original$\uparrow$ & Reversed$\uparrow$ & Combined$\uparrow$ & Original$\uparrow$ & Reversed$\uparrow$ & Combined$\uparrow$ \\\midrule
Camera original & \metricstd{86.38}{0.86} & \metricstd{75.66}{0.68} & \metricstd{81.02}{0.15} & \metricstd{83.15}{0.13} & \metricstd{76.42}{0.43} & \metricstd{79.78}{0.28} \\
35K factual-pair control & \metricstd{87.60}{0.43} & \metricstd{77.71}{0.13} & \metricstd{82.66}{0.17} & \metricstd{84.95}{1.13} & \metricstd{78.51}{0.17} & \metricstd{81.73}{0.51} \\
\rowcolor{storymotionblue}
+ DPA & \textbf{\metricstd{89.66}{0.67}} & \textbf{\metricstd{79.00}{0.22}} & \textbf{\metricstd{84.33}{0.28}} & \textbf{\metricstd{86.63}{0.77}} & \textbf{\metricstd{79.58}{0.50}} & \textbf{\metricstd{83.11}{0.14}} \\
\bottomrule\end{tabular*}\endgroup

\end{table}

\paragraph{Paired sampling and scoring.}
\acceptededitorial{Each prompt arm contains the same 3,194 clips with active reference translation, giving 6,388 clip--prompt cases per task. \acceptededitorial{Original and reversed prompts are processed by the same text encoder; human latents, camera noise and valid-length masks are shared within each pair.} Given-human uses reference motion; joint generation samples the human first.}

Direction accuracy measures the sign of each reference-active translation axis at fixed reference time positions. Each clip first averages sign correctness over those positions and axes, then clips receive equal weight. Reversed targets negate the corresponding reference signs. Reference-inactive positions are excluded; generated still predictions at active positions count as errors. Combined accuracy averages the two prompt arms.

\FloatBarrier
\section{Translation-Intensity Control}

\label{supp:results}\label{supp:intensity}
\acceptededitorial{For the endpoint change from $a=1$ to $1.5$, path length is nondecreasing for 97.31\% of given-human test outputs and 97.01\% of joint test outputs.}
\section{User Study}
\label{supp:user-study}
\begin{figure}[!htbp]
\centering
\includegraphics[width=\linewidth]{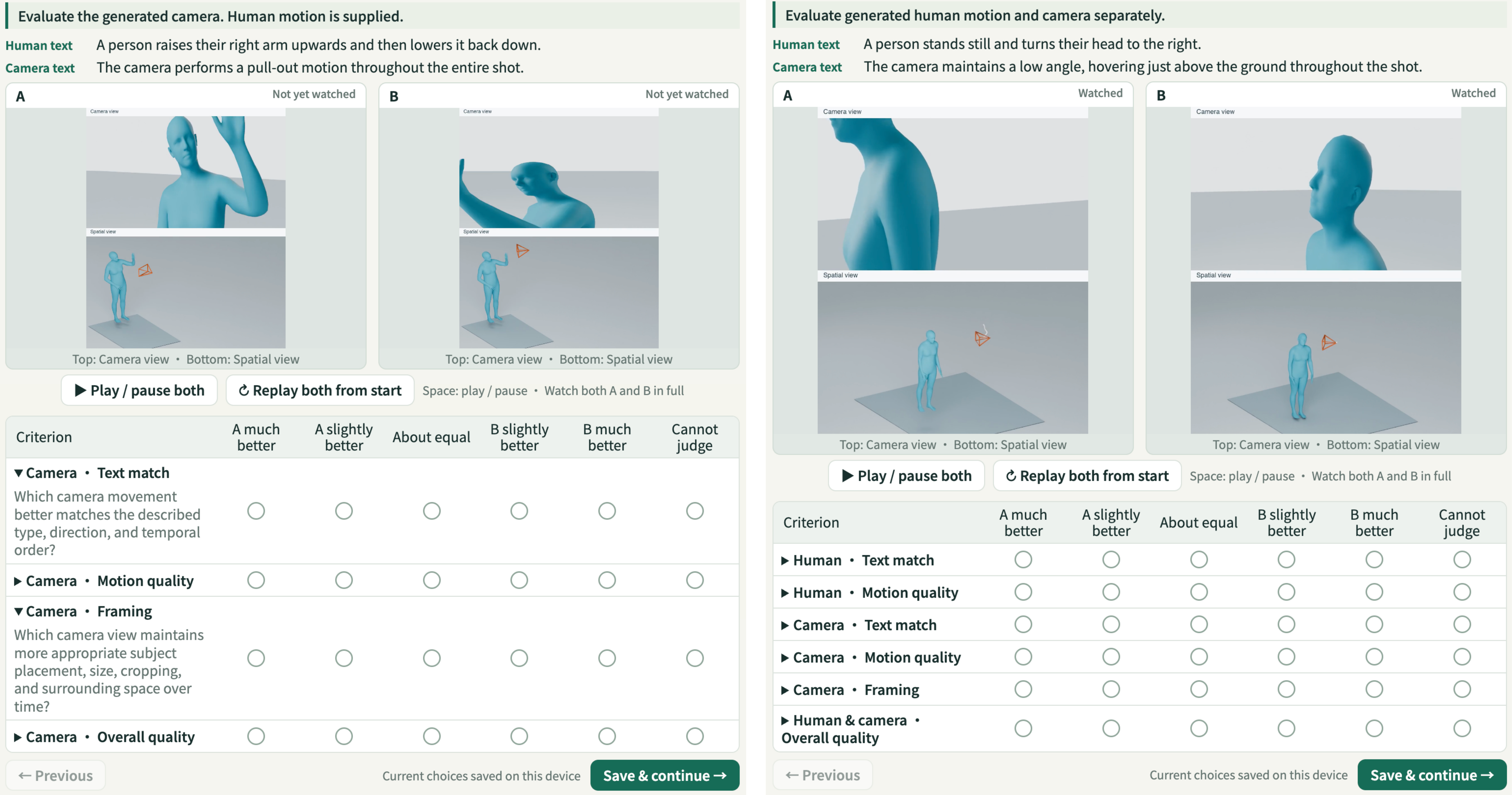}\\[3pt]
\makebox[.49\linewidth]{\textbf{(a) Given-human camera generation}}\hfill
\makebox[.49\linewidth]{\textbf{(b) Joint human--camera generation}}
\caption{\textbf{User-study interface.} Anonymized A/B videos show the generated camera view above an external spatial view. Participants compare text alignment, motion quality, framing and task-specific overall quality. The joint task additionally evaluates human motion.}
\label{fig:user-study-interface}
\end{figure}

\paragraph{Study design.}
\acceptededitorial{The anonymous study follows the task allocation in Section~\ref{sec:user-study} and includes baseline--baseline comparisons. Each questionnaire receives a randomized, balanced assignment with equal target exposure per method pair, balanced task order, and randomized left--right presentation and trial order. Figure~\ref{fig:user-study-interface} shows synchronized camera and spatial views.}

\paragraph{Evaluation criteria.}
Both tasks assess camera-text alignment, camera motion quality and framing. Given-human trials additionally ask for overall camera quality; joint trials add human-text alignment, human motion quality and overall human--camera quality. Each criterion uses five responses from ``A much better'' to ``B much better,'' including a tie, plus ``Cannot judge.'' Overall quality is judged directly. Both videos must reach 90\% playback coverage before a trial can be saved.

\paragraph{Analysis.}
\acceptededitorial{For each clip, method pair and criterion, we compute preference as \((W + 0.5T)/(W + T + L)\), where \(W\), \(T\) and \(L\) count wins, ties and losses. ``Cannot judge'' responses are counted separately. Within each task and criterion, pairwise scores average equally over clips, and method-level scores average equally over opponents. Overall quality is the primary outcome for each task. We estimate 95\% confidence intervals using 10,000 participant-cluster bootstrap resamples with the stimulus set fixed.}

\paragraph{\acceptededitorial{Results.}}
\acceptededitorial{The 29 complete questionnaires comprise 580 given-human and 522 joint trials, covering all 120/60 clip--pair comparisons. Of 5,452 criterion responses, 5,238 are judgeable and 214 are ``Cannot judge.''}

\begin{figure}[!htbp]
\centering
\includegraphics[width=\textwidth]{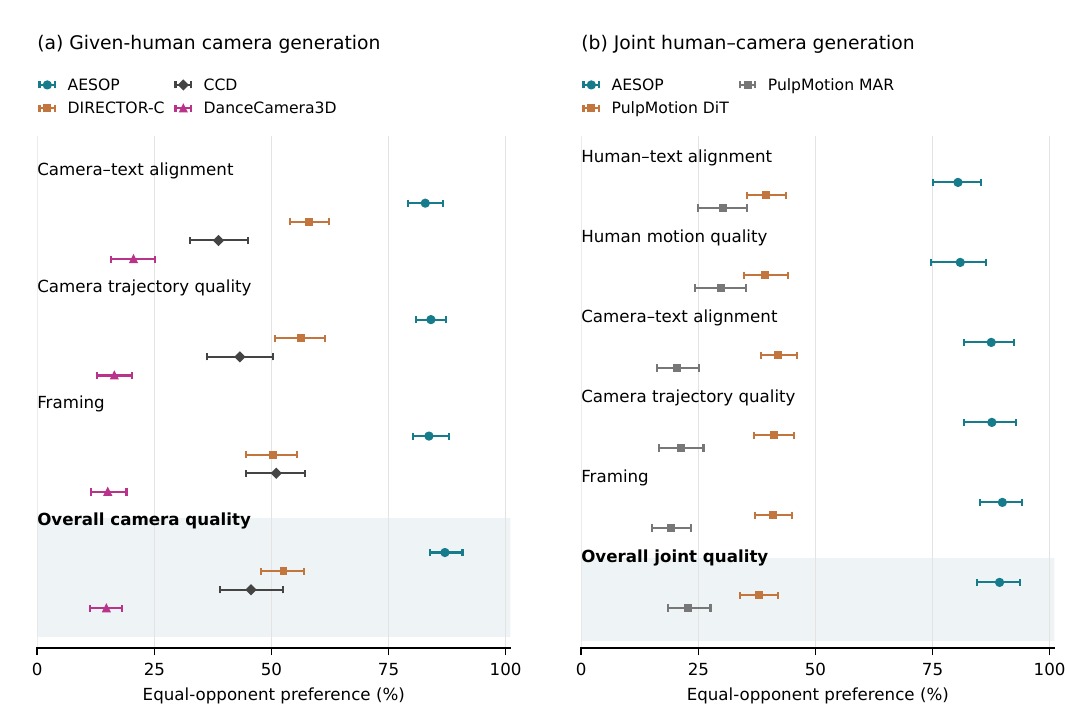}
\caption{\acceptededitorial{\textbf{User-study preferences across all criteria.} Points show clip-balanced, equal-opponent preference scores; bars show 95\% participant-cluster bootstrap intervals. The vertical dashed line marks 50\% preference. Shading highlights the separately rated overall-quality endpoints.}}
\label{fig:user-study-results}
\end{figure}

\acceptededitorial{Figure~\ref{fig:user-study-results} reports all task-specific criteria. AESOP has the highest preference on each criterion: joint human-text alignment and motion-quality scores are 80.4\% and 80.9\%, while camera alignment, motion-quality and framing scores range from 87.5\% to 89.9\%. \acceptededitorial{Section~\ref{sec:user-study} and Table~\ref{tab:user-study-overall} report overall preference.}}

\FloatBarrier
\section{\acceptededitorial{Qualitative results on external motions and prompts}}
\label{supp:ood-humanml3d}
\acceptededitorial{We demonstrate the same PulpMotion-trained AESOP model in three settings: camera generation from HumanML3D motions, joint generation from HumanML3D action descriptions, and camera generation from externally synthesized human motions.}

\begin{figure}[!htbp]
\centering
\includegraphics[width=0.90\textwidth]{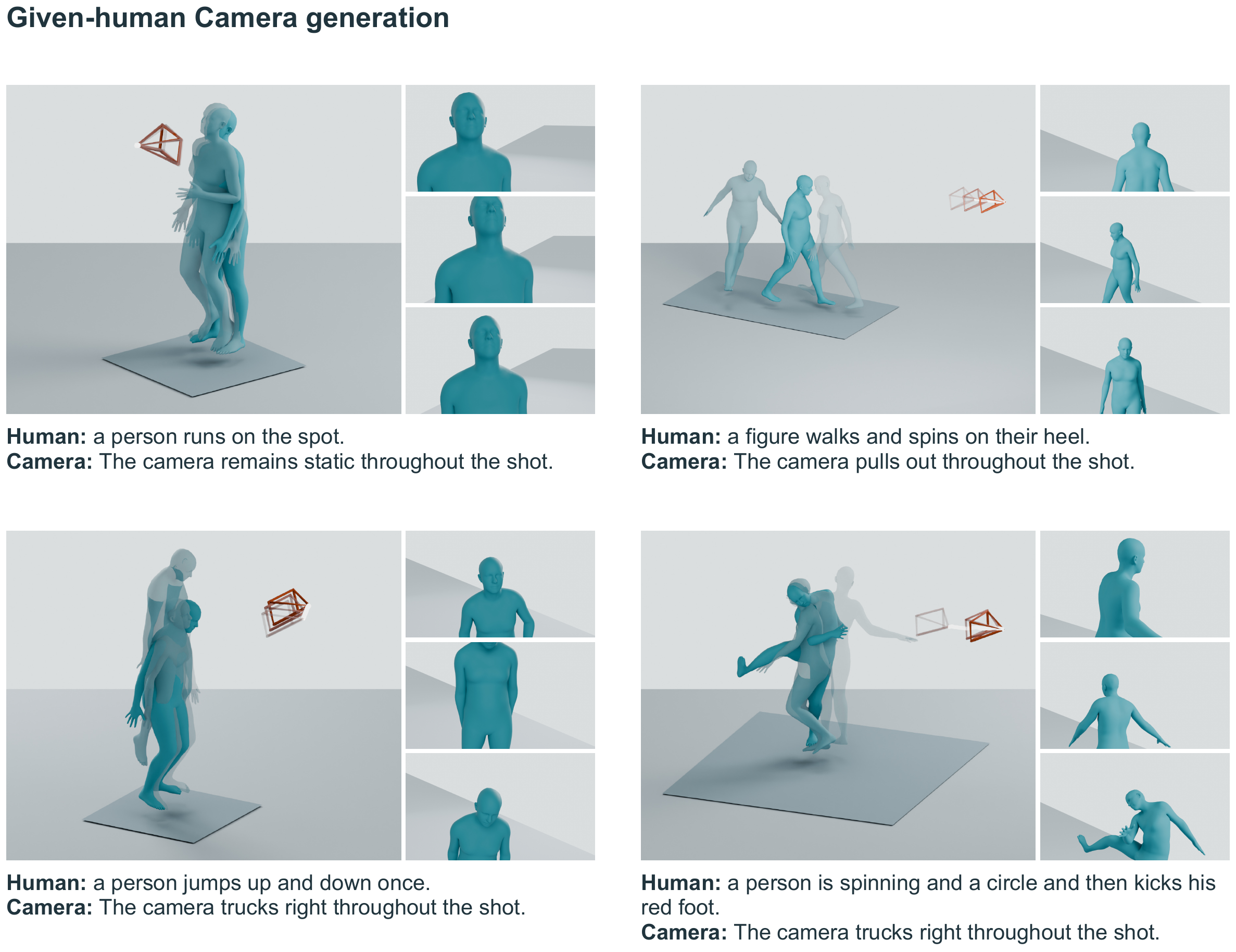}
\caption{\textbf{Camera generation conditioned on HumanML3D motions.} Each example shows a global view and three successive camera views; lighter poses indicate earlier frames. Global views are fitted separately to each scene.}
\label{fig:ood-given}
\end{figure}

\paragraph{Ground-truth human motions.}
We adapt HumanML3D~\citep{guo2022humanml3d} motions to the human representation and generate cameras conditioned on these motions and camera prompts. Figure~\ref{fig:ood-given} shows a static camera observing running in place, a pull-out widening the view of a walking and turning figure, and trucking accompanying jumping and spinning.

\begin{figure}[!htbp]
\centering
\includegraphics[width=0.90\textwidth]{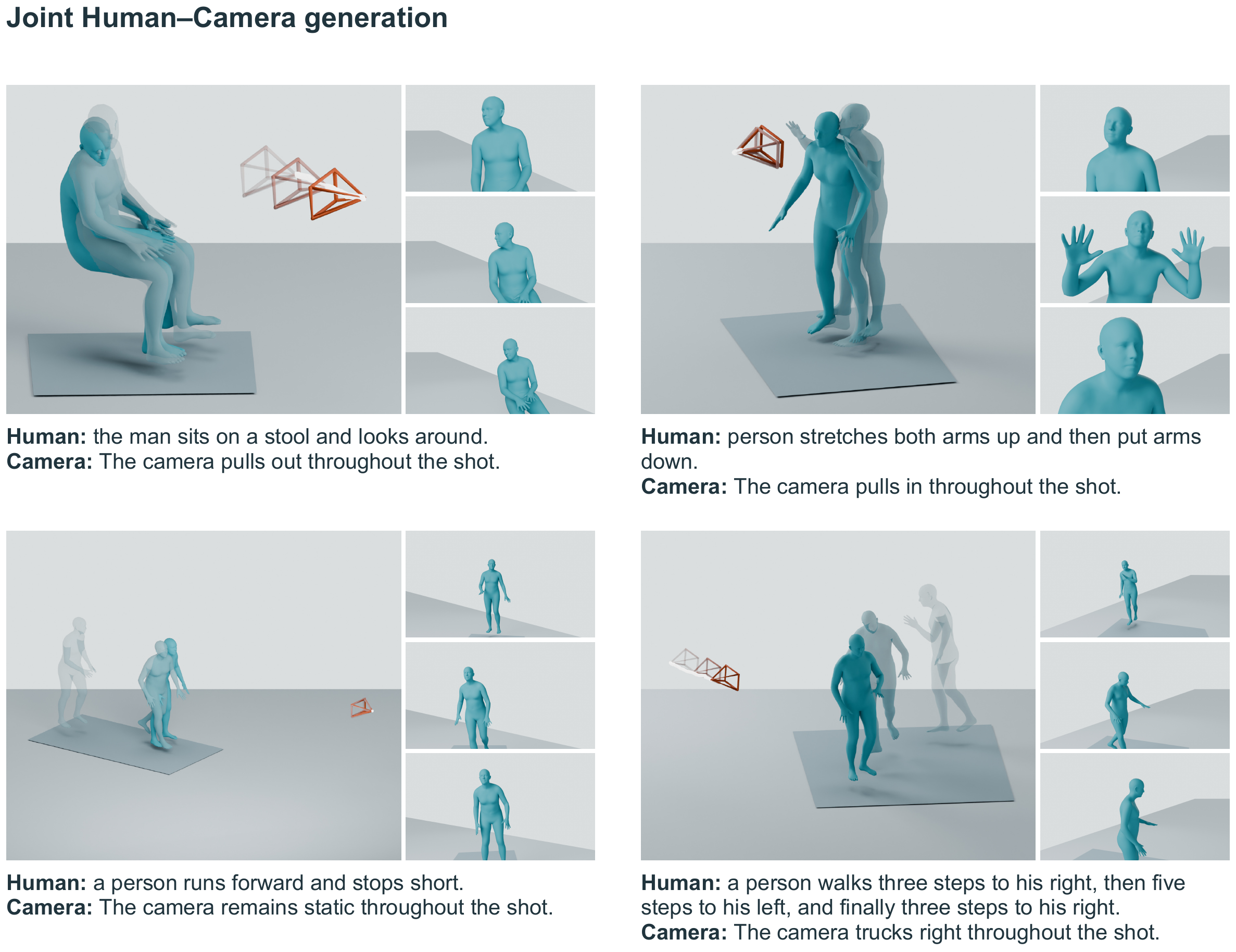}
\caption{\textbf{Joint generation from HumanML3D action descriptions.} Each example shows generated human motion and camera trajectory in a global view and three successive camera views.}
\label{fig:ood-joint}
\end{figure}

\paragraph{Joint generation from action descriptions.}
We use ground-truth HumanML3D action descriptions to generate both human motion and camera trajectory (Figure~\ref{fig:ood-joint}). A pull-in tightens the view of arm stretching, while a static camera observes an approaching runner.

\begin{figure}[!htbp]
\centering
\includegraphics[width=0.90\textwidth]{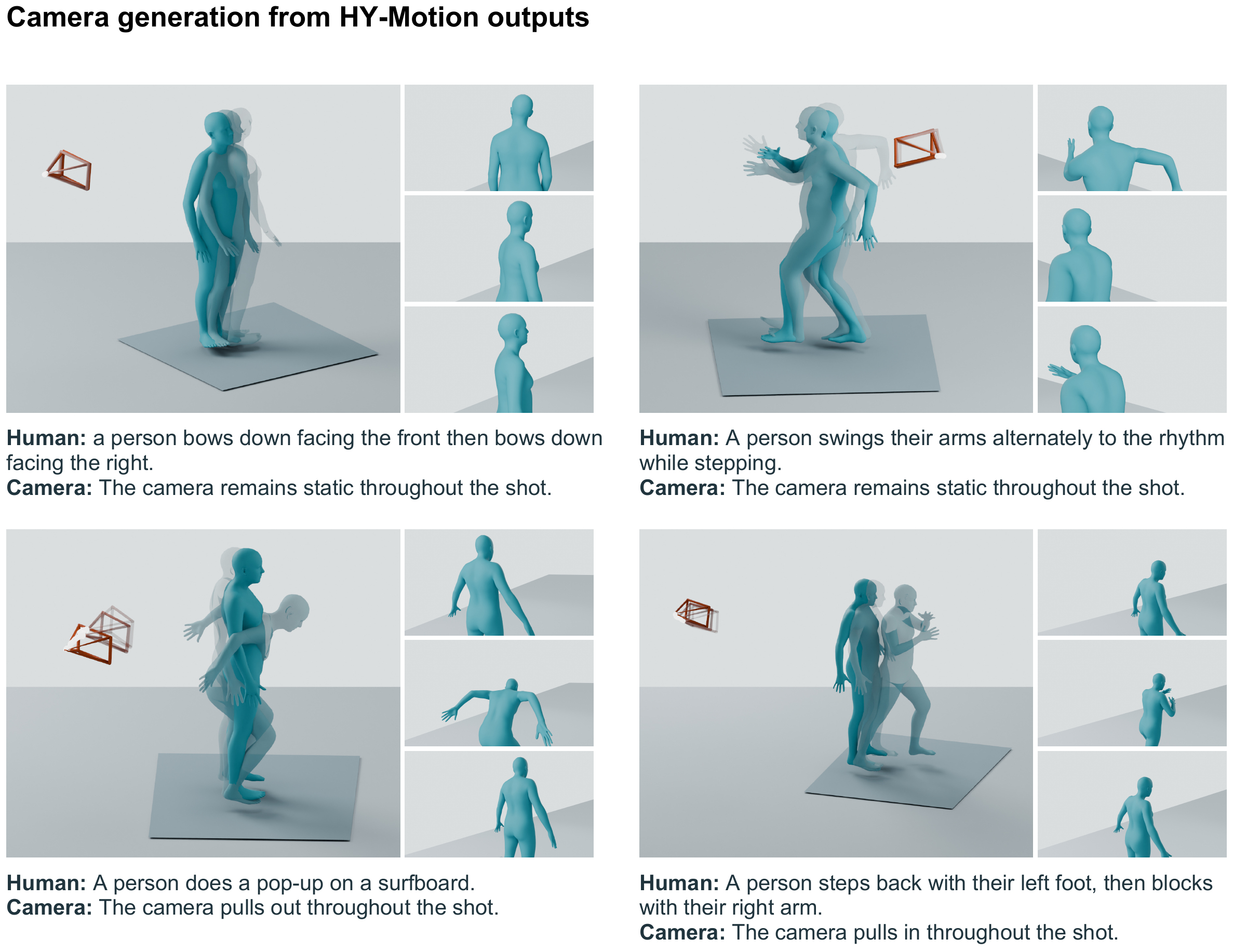}
\caption{\textbf{Camera generation conditioned on HY-Motion outputs.} Four synthesized human motions are paired with static, pull-out and pull-in camera prompts. Each example includes a global view and three successive camera views.}
\label{fig:ood-synthesized}
\end{figure}

\paragraph{Synthesized human motions.}
We condition the same camera generator on motions synthesized by HY-Motion. Figure~\ref{fig:ood-synthesized} shows static views for bowing and rhythmic stepping, a pull-out for a surfboard pop-up, and a pull-in for stepping back and blocking.

\FloatBarrier

\clearpage
\section{More Qualitative Results}
\label{supp:more-results}
Figures~\ref{fig:more-given} and~\ref{fig:more-joint} show two additional examples for each task. Each method shows a global view and three camera projections in temporal order from top to bottom.
\begingroup
\setlength{\intextsep}{6pt}
\begin{figure}[H]
\centering
\includegraphics[width=0.90\textwidth]{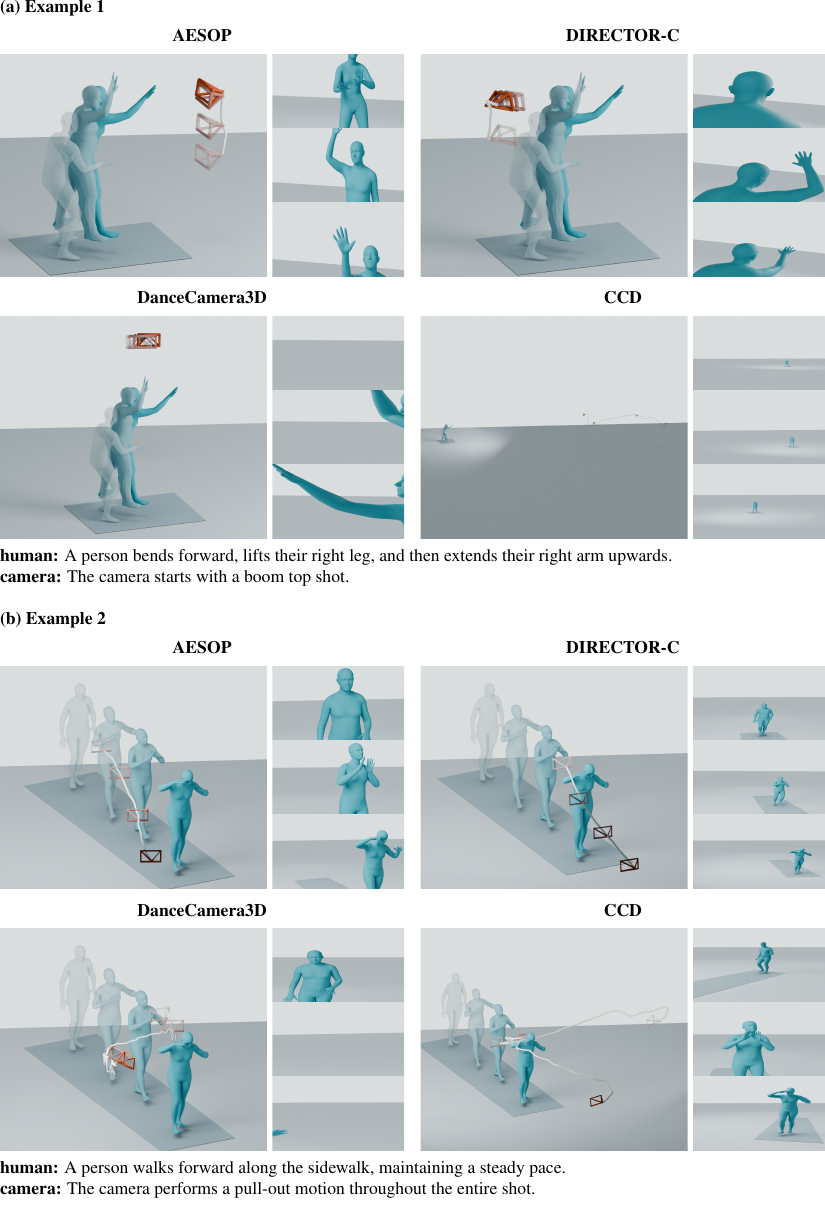}
\caption{\textbf{Additional given-human camera generation examples.} Global views are fitted separately to each trajectory.}
\label{fig:more-given}
\end{figure}

\clearpage
\begin{figure}[H]
\centering
\includegraphics[width=0.90\textwidth]{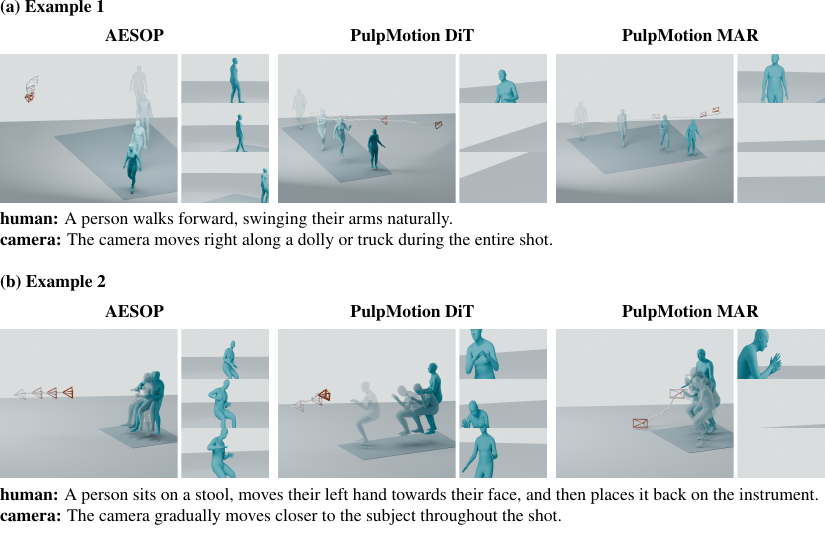}
\caption{\textbf{Additional joint human--camera generation examples.} Global views are fitted separately to each trajectory.}
\label{fig:more-joint}
\end{figure}

\endgroup
\FloatBarrier

\FloatBarrier

\end{document}